\documentclass{article} % For LaTeX2e
\usepackage{colm2024_conference,times}

\usepackage{booktabs}
\usepackage{graphicx}
\usepackage{enumitem}
\usepackage{wrapfig}
\usepackage{algorithm}
\usepackage{algpseudocode}

\usepackage{amsmath,amsfonts,bm}

\def\eqref#1{equation~\ref{#1}}
\def\1{\bm{1}}

\DeclareMathAlphabet{\mathsfit}{\encodingdefault}{\sfdefault}{m}{sl}
\SetMathAlphabet{\mathsfit}{bold}{\encodingdefault}{\sfdefault}{bx}{n}

\usepackage{microtype}
\usepackage{amsmath,amssymb}
\usepackage{caption}      % for \captionof
\usepackage{subcaption}   % optional, but plays nicely with side-by-side layouts
\usepackage{hyperref}
\usepackage{multirow}
\usepackage{colortbl,xcolor}
\usepackage{siunitx}
\usepackage[most]{tcolorbox}
\tcbset{
  colback=gray!10,
  colframe=gray!40,
  boxrule=0.3pt,
  arc=2mm,
  left=8pt,right=8pt,top=8pt,bottom=8pt
}
\newtcolorbox[auto counter, number within=section]{promptbox}[2][]{ coltitle=black, fonttitle=\bfseries, enhanced, breakable, #1}

\newcommand{\equalcontrib}{\textsuperscript{*}}   % equal contribution mark
\newcommand{\corrauth}{\textsuperscript{\dag}}    % corresponding author mark
\newcommand{\methodname}{\textsc{VideoAgentTrek}}
\newcommand{\screenfilter}{\textsc{ScreenFilter}}
\newcommand{\videoaction}{\textsc{Video2Action}}

\definecolor{LightBlue}{RGB}{230,245,255}

\title{VideoAgentTrek: Computer Use Pretraining from Unlabeled Videos}

\author{
Dunjie Lu\textsuperscript{1,2}\equalcontrib,
Yiheng Xu\textsuperscript{1,2}\equalcontrib,
Junli Wang\textsuperscript{2}\equalcontrib,
Haoyuan Wu\textsuperscript{1},
Xinyuan Wang\textsuperscript{1},
Zekun Wang\textsuperscript{2},\\
Junlin Yang\textsuperscript{1},
Hongjin Su\textsuperscript{1},
Jixuan Chen\textsuperscript{1},
Junda Chen\textsuperscript{1},
Yuchen Mao\textsuperscript{1},\\
Jingren Zhou\textsuperscript{2},
Junyang Lin\textsuperscript{2},
Binyuan Hui\textsuperscript{2}\corrauth,
Tao Yu\textsuperscript{1}\corrauth \\
\textsuperscript{1}The University of Hong Kong \qquad
\textsuperscript{2}Qwen Team, Alibaba Group
}

\begin{document}

\maketitle

% footnotes for the title page
\begingroup
\renewcommand\thefootnote{\fnsymbol{footnote}} % * † ‡ ...
\footnotetext[1]{Equal contribution.\enspace \textsuperscript{\dag}~Corresponding authors.}
\endgroup

\begin{abstract}
% Training computer-use agents requires massive amounts of GUI interaction data, but manually annotating action trajectories at scale is prohibitively expensive. We present \methodname, a scalable pipeline that automatically mines training data from publicly available screen-recorded videos, eliminating the need for manual annotation. Our approach addresses a key challenge: raw videos contain implicit demonstrations but lack explicit action labels. To solve this, we develop \videoaction{}, an inverse dynamics module (IDM) with two components: (1) a video grounding model that detects and localizes GUI actions with precise temporal boundaries, and (2) an action-content recognizer that extracts structured parameters like click coordinates and typed text. Applied to 39,000 YouTube tutorial videos, our pipeline generates 1.52 million interaction steps. We leverage this data through continued pretraining followed by supervised fine-tuning. On OSWorld-Verified, our approach improves task success rates from 9.3\% (SFT-only baseline) to 15.8\%, a 70\% relative improvement. On AgentNetBench, step accuracy increases from 64.1\% to 69.3\%. Our results demonstrate that passive internet videos can be transformed into high-quality supervision for computer-use agents, providing a scalable alternative to expensive manual annotation.
Training computer-use agents requires massive amounts of GUI interaction data, but manually annotating action trajectories at scale is prohibitively expensive. We present \methodname, a scalable pipeline that automatically mines training data from publicly available screen-recorded videos at web scale, eliminating the need for manual annotation. Our approach addresses a key challenge: raw videos contain implicit demonstrations but lack explicit action labels. To solve this, we develop \videoaction{}, an inverse dynamics module (IDM) with two components: (1) a video grounding model that detects and localizes GUI actions with precise temporal boundaries and context, and (2) an action-content recognizer that extracts structured parameters like click coordinates and typed text with high fidelity. Applied to 39,000 YouTube tutorial videos, our pipeline generates 1.52 million interaction steps automatically. We leverage this data through continued pretraining followed by supervised fine-tuning. On OSWorld-Verified, our approach improves task success rates from 9.3\% (SFT-only baseline) to 15.8\%, a 70\% relative improvement. On AgentNetBench, step accuracy increases from 64.1\% to 69.3\%. Our results demonstrate that passive internet videos can be transformed into high-quality supervision for computer-use agents, providing a scalable alternative to expensive manual~annotation.

\end{abstract}

\section{Introduction}
Teaching machines to use computers like humans do (clicking buttons, typing text, navigating interfaces) represents a fundamental challenge in AI. While recent advances in vision-language models have made computer-use agents increasingly feasible~\citep{bai2025qwen25vltechnicalreport, qin2025uitarspioneeringautomatedgui,kimiteam2025kimivltechnicalreport,wang2025opencuaopenfoundationscomputeruse}, their development remains bottlenecked by data availability. Training these agents requires extensive trajectories that precisely document GUI interactions: screenshots paired with exact action parameters like click coordinates $(x, y)$ and typed strings. However, creating such datasets through manual annotation is extraordinarily expensive, making it impractical to achieve the scale necessary for robust generalization across diverse applications and operating systems.

% Meanwhile, the internet hosts millions of screen-recorded tutorials where humans naturally demonstrate computer use, from Excel tutorials to software walkthroughs. These videos implicitly contain the exact supervision we need: they show where users click, what they type, and how interfaces respond. Yet this vast resource remains untapped because videos lack the structured action labels required for training. The cursor movements are visible but not tracked; the typed text appears but isn't extracted; the timing of actions is implicit but not annotated. We can learn to automatically extract structured action trajectories from raw videos by training specialized models to detect \emph{when} actions occur and infer \emph{what} their parameters are, effectively converting passive recordings into active training data.

Meanwhile, the internet hosts millions of screen-recorded tutorials where humans demonstrate computer use, from Excel tutorials to software walkthroughs. These videos implicitly contain the supervision we need: they show where users click, what they type, and how interfaces respond. Yet this resource remains untapped because videos lack the structured action labels required for training. The cursor movements are visible but not tracked; the typed text appears but isn't extracted; the timing of actions is implicit but not annotated. We can learn to automatically extract structured action trajectories from raw videos by training specialized models to detect \emph{when} actions occur and infer \emph{what} their parameters are, effectively converting passive recordings into active training data.

We introduce \textbf{\methodname}, a scalable pipeline that mines computer-use trajectories from publicly available unlabeled videos without manual annotation. Our approach employs \videoaction{}, an inverse dynamics module (IDM) with two stages: First, an action event detection model performs dense event detection, identifying action types and their precise temporal boundaries (e.g., \texttt{click} at $[1.5, 2.0]$s, \texttt{type} at $[3.5, 5.5]$s). Second, the action parameterization model, an action-content recognizer, analyzes these localized segments to extract structured parameters (pointer coordinates for clicks, literal text for typing), yielding complete $(screenshot, action, parameters)$ trajectories suitable for training.

\methodname{} enables large-scale computer-use pretraining with unlabeled web videos.
From 39,000 YouTube videos, we automatically extract 1.52 million interaction steps.
This represents not just more data, but more diverse data: the trajectories span hundreds of applications across Windows, macOS, and web platforms, capturing interaction patterns that would be infeasible to annotate manually.

We validate \methodname{} with a two-stage training recipe: continued pretraining on the mined trajectories followed by supervised fine-tuning on a curated dataset.
This combination leverages the broad coverage from videos to learn fundamental GUI interaction patterns, while supervised fine-tuning sharpens task-specific performance.
Our models achieve 15.8\% task success on OSWorld-Verified compared to 9.3\% for baselines, a 70\% relative improvement.
The gains are particularly pronounced in online environments where robustness to visual variation matters most. We summarize our main contributions and findings below:
\begin{itemize}[leftmargin=*]
  \item We propose \methodname, an unsupervised approach to training computer-use agents that automatically converts screen-recorded videos into structured training data through learned inverse dynamics, thereby eliminating the need for manual annotation.
  \item Our \videoaction{} module implements inverse dynamics, combining action event detection with millisecond-precision temporal localization and action parameter extraction. It enables accurate reconstruction of GUI interactions (clicks, typing,...) from raw video without ground-truth labels.
  \item Experiments demonstrate that our approach achieves 15.8\% task success on OSWorld-Verified compared to 9.3\% for SFT-only baselines (70\% relative improvement), and improves step accuracy on AgentNetBench from 64.1\% to 69.3\%, validating that passive internet videos can provide effective supervision at scale.
  \item We provide a reproducible pipeline and training methodology that enables researchers to leverage publicly available screen recordings for computer-use agent training. To facilitate future research, we release \screenfilter{} for efficient GUI filtering and \videoaction{} for action extraction as open-source tools.
\end{itemize}

\begin{figure}[t]
\vspace{-5mm}
  \centering
  \includegraphics[width=1.0\linewidth]{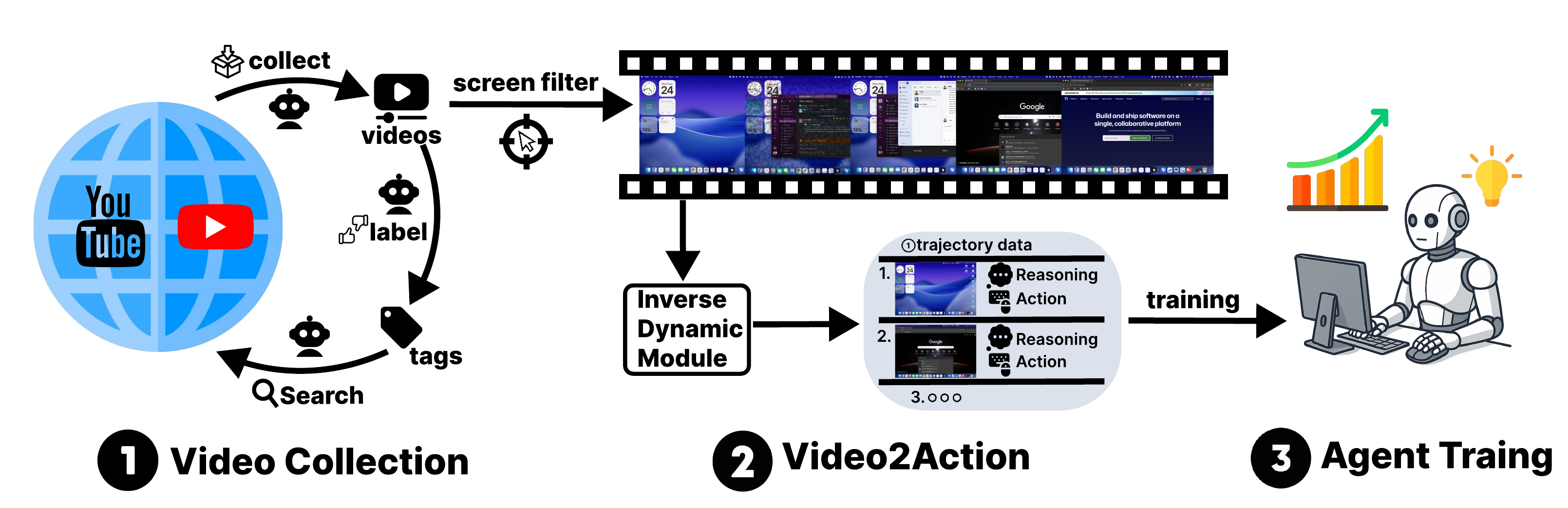}
  \caption{Overview of \methodname. (1) \textbf{Video Collection}: crawl screen-recorded tutorials and filter GUI footage with \screenfilter{}. (2) \textbf{Video2Action}: an inverse dynamics module that first performs dense action-event detection to localize clips and assign action types, then \emph{action parameterization} (e.g., click coordinates, typed text) to yield structured $(\text{screenshot}, \text{action}, \text{parameters})$ trajectories. (3) \textbf{Agent Training}: use the mined trajectories for continued pretraining and supervised finetuning of computer-use agents.}
  \label{fig:mainfigure}
\end{figure}

\section{VideoAgentTrek}\label{sec:videoagenttrek}

We introduce \textbf{\methodname}, a video-driven pipeline that turns web tutorials into training supervision for computer-use agents. Each trajectory is a sequence $\mathcal{R}=\{(I_k, r_k, a_k, \pi_k)\}_{k=1}^{K}$ following~\cite{react}, where $I_k$ is a representative screenshot, $r_k$ is a brief inner monologue, $a_k\in\mathcal{A}$ is the action type (e.g. click, type), and $\pi_k$ is the action content (e.g., pointer $(x,y)$ or typed text). The pipeline has three parts:

\begin{itemize}[leftmargin=*]
  \item \textbf{Video collection and preprocessing.} We crawl tutorial videos with seeded queries and tag expansion, apply human-in-the-loop screening, and use cursor-based filtering to retain screen segments with GUI interactions (Section~\ref{sec:method:videocollectionandpreprocessing}).
  \item \textbf{\videoaction{}.} From raw video, we recover stepwise supervision without manual labels: (i) dense event detection produces typed segments with tight start/end times; (ii) action identification infers parameters $\pi_k$ (e.g., click coordinates, typed strings); and (iii) a short inner monologue $r_k$ makes the intent explicit. Assembling these per-clip steps yields ReAct tuples for training (Section~\ref{sec:method:inversedynamicmethod}).
  \item \textbf{Agent training.} We combine large-scale agentic data produced by the method with human demonstrations and targeted GUI grounding pairs, and train an end-to-end agent in two stages: interleaved video–text pretraining followed by instruction-style finetuning (Section~\ref{sec:method:building_computer_use_agent}).
\end{itemize}

This structure scales supervision to web-scale while preserving the stepwise semantics needed for robust computer-use policies.

% \section{Data Collection}\label{sec:data collection}
\begin{figure}[t]
    % \vspace{-5mm}
  \centering
  \includegraphics[width=0.8\linewidth]{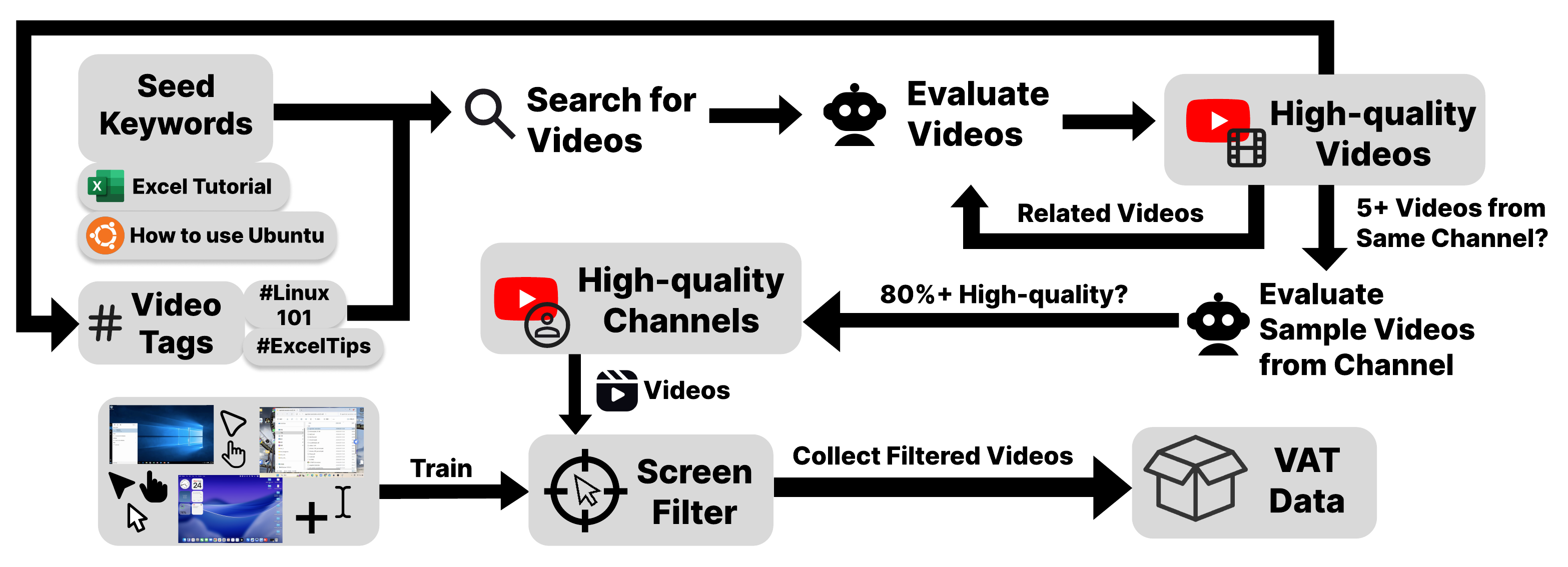}
  \caption{\textbf{Video candidate auto-discovery.} From seed keywords and tags, we search and evaluate videos, expand to related videos and high-quality channels ($\geq$80\% pass), and iteratively collect GUI-containing videos for VAT.}
  \label{fig:video_collection_pipeline}
  % \vspace{-5mm}
\end{figure}

\subsection{Video collection and preprocessing}
\label{sec:method:videocollectionandpreprocessing}

\subsubsection{Video Candidate Auto-Discovery}

We employ a scalable pipeline for video collection that leverages channel coherence---the observation that YouTube channels typically maintain consistent content types and quality.
Starting from seed keywords such as ``Excel tutorial" and ``How to use Windows", we validate initial results and extend to entire channels when sampling indicates high quality (i.e., when $\geq$ 80\% of samples meet our criteria).
This channel-based expansion enables efficient scaling: validated channels become trusted sources, while their tags and metadata enable iterative discovery.

When we identify high-quality channels through seed validation, we include all their videos as candidates rather than individually vetting each one.
This approach deliberately optimizes recall over precision, as the subsequent \screenfilter{} stage ensures final data quality.
The channel coherence property---where content creators typically focus on consistent topics---makes this expansion particularly effective.

Through iterative rounds of keyword search, channel expansion, and tag extraction, we transform a small set of manually validated seeds into 55{,}000 candidate videos ($\sim$10{,}000 hours).
This process requires minimal human oversight: initial quality validation on seed videos and periodic verification of expansion effectiveness.
The resulting candidate pool intentionally includes some non-GUI content (presentations, tutorials with mixed content), which our filtering stage handles efficiently.

\subsubsection{Video Preprocessing with \screenfilter{}}

Although keyword-based searches typically retrieve relevant computer operations, they also include non-interactive segments, such as explanatory sections where the presenter uses PowerPoint or other presentation tools. Additionally, some of the videos retrieved through this method may not meet the standards for GUI interaction content.

To address this, we developed \screenfilter{}, a lightweight cursor detection model upon YOLOv8x ~\citep{reis2024realtimeflyingobjectdetection} to efficiently extract video segments that focus exclusively on GUI interactions. Using the detection results, we retain video segments where at least 80\% of the frames contain a cursor for 6 seconds or more, with a 2-second merge gap for temporal smoothing. When applied to our corpus, \screenfilter{} successfully extracts 7,377 hours of verified GUI interactions from 10,000 hours of raw video. \screenfilter{}’s details are in Appendix~\ref{app:screenfilterdetails}.

\subsubsection{VideoAgentTrek Data Analysis}\label{sec:analysis:videoagenttrekdataanalysis}

\noindent\textbf{Quality and relevance.}
We collected 55k screen-capture videos (about 10{,}000 hours) from 50{+} channels. The corpus is predominantly clear (about 97\% are 720p or higher) and most clips are minutes long, yielding sustained, readable interactions suitable for our pipeline (Table~\ref{tab:vatt-res}). A lightweight title/description audit groups videos into tutorials, background pieces, tech talks, and unrelated; tutorials dominate (69.6\%), with the remainder used mainly for tag mining or removed during filtering (Table~\ref{tab:l1_content_form_table}). Together, these checks indicate that the collected data are both visually clean and topically aligned with computer-use supervision.

% \begin{wrapfigure}{r}{0.3\linewidth}
\begin{wrapfigure}[7]{r}{0.3\linewidth} 
    \vspace{-11mm}
    \centering
    \includegraphics[width=0.7\linewidth]{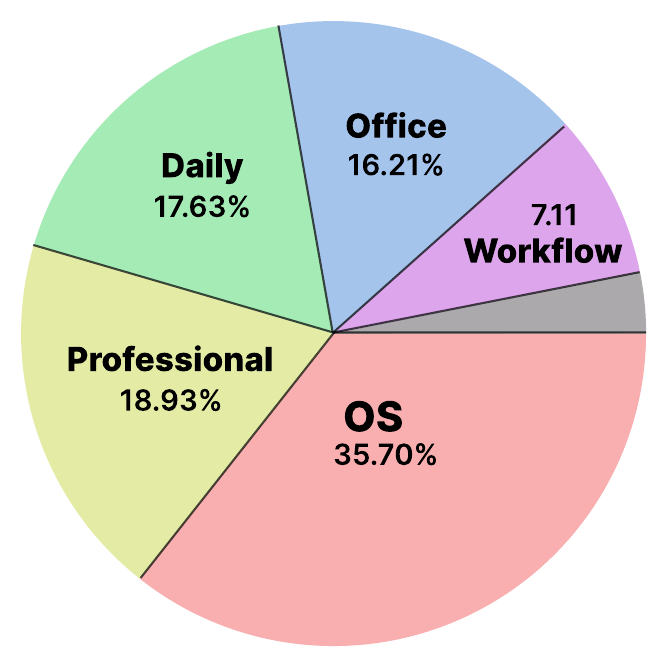}
    \vspace{-2mm}
    \caption{Domain distribution.}
    \label{fig:data_classification}
    % \vspace{-10mm}
\end{wrapfigure}

\noindent\textbf{Data classification.}
We label each video as daily, office, workflow, professional, operating-system (OS), or other using a lightweight GPT-4.1 pass over the title and a short transcript snippet. The distribution (Figure~\ref{fig:data_classification}) is skewed toward OS-level operations (\(\sim\)36\%), followed by professional (\(\sim\)19\%), daily (\(\sim\)18\%), and office (\(\sim\)16\%); workflow is smaller (\(\sim\)7\%) with a small remainder labeled as other (\(\sim\)4\%). This indicates broad coverage with a bias toward system and professional use cases.

\subsection{\videoaction{}: Inverse Dynamics Module}
\label{sec:method:inversedynamicmethod}

We develop \videoaction{}, an inverse dynamics module (IDM) that extracts structured action supervision from unlabeled GUI videos.
Following insights from robotics where inverse dynamics recovers actions from observations~\citep{nguyen2010using}, \videoaction{} detects GUI events (clicks, drags, scrolls, typing) and infers their parameters directly from pixel changes.
This yields training-ready (screenshot, action) pairs without manual annotation, forming the second core component of our VideoAgentTrek toolkit.

\begin{figure}[t]
    % \vspace{-5mm}
  \centering
  \includegraphics[width=0.8\linewidth]{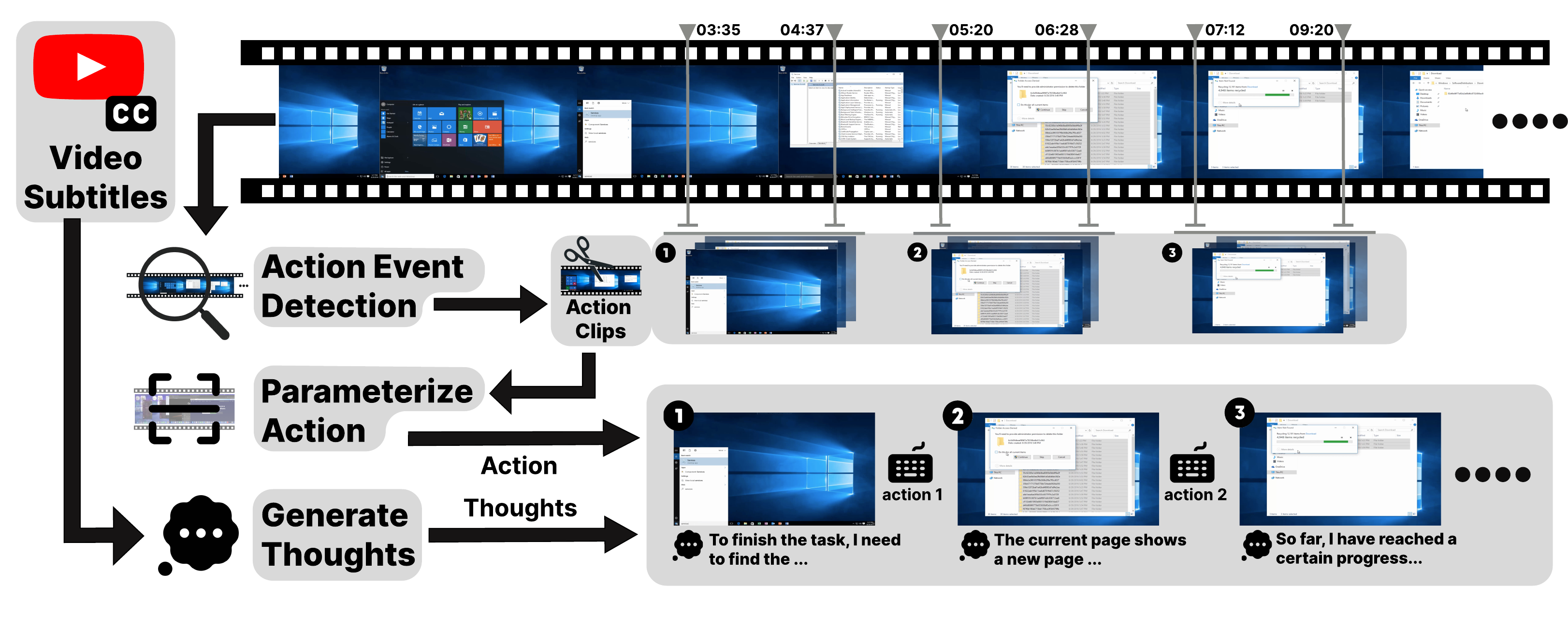}
  \caption{\textbf{Overview of \videoaction{}}: Given a screen-capture video (with optional subtitles), the module (1) detects GUI action events and segments clips, (2) parameterizes each action (type and arguments), and (3) generates step-level thoughts, yielding training-ready sequences of \{action clip, action, thought\}}
  \label{fig:inverse_dynamic_method}
  \vspace{-5mm}
\end{figure}

\subsubsection{Action Event Detection}\label{sec:method:denseeventdetection}
\textbf{Task.}
Given an unlabeled screen-capture video $v$ (length $T$), perform prompt-free, dense event detection: predict a set of typed GUI interactions with tight temporal bounds,
\[
f_\theta(v)\ \rightarrow\ \mathcal{S}=\{(a_k,t_k^{\mathrm{s}},t_k^{\mathrm{e}})\}_{k=1}^{K},\quad a_k\in\mathcal{A},\ 0\le t_k^{\mathrm{s}}<t_k^{\mathrm{e}}\le T.
\]
Unlike query-based setups, our input contains only $v$; the output is a multi-event set with both action types and start/end timestamps.

\textbf{Approach.}
We equip a VLM with video grounding so that, given a clip, it emits a sequence of $(a_k, t^{\mathrm{s}}_k, t^{\mathrm{e}}_k)$ for all GUI actions, reframing keyframe detection as multi-class temporal event detection with tight bounds. 
\textbf{(1) Training data}: We utilize the annotation tool provided by OpenCUA~\citep{wang2025opencuaopenfoundationscomputeruse} to obtain synchronized screen videos and timestamped GUI interactions (mouse/keyboard events). These raw demonstration logs are then used to create temporal-grounding supervision, allowing precise event detection without manual annotation.
% We build on OpenCUA \citep{wang2025opencuaopenfoundationscomputeruse}, which provides an annotation tool that records complete human computer-use demonstrations with synchronized screen video and precisely timestamped GUI interactions (mouse/keyboard events and UI context). We use the raw demonstration logs as our source and convert them into temporal-grounding supervision. This produces clean, prompt-free labels for dense event detection while preserving the original timing fidelity needed for tight temporal bounds.
\textbf{(2) Model training}: We leverage Qwen2.5-VL~\citep{bai2025qwen25vltechnicalreport} as the base model, benefiting from its multimodal understanding and fine-grained spatiotemporal capabilities. We perform full-parameter supervised fine-tuning on the Qwen2.5-VL-7B-Instruct model to enable it to generate ordered, typed event spans directly from raw video clips.
% We choose Qwen2.5-VL \citep{bai2025qwen25vltechnicalreport} as the foundation, leveraging its long-context multimodal understanding and fine-grained spatiotemporal reasoning capabilities. We perform full-parameter supervised fine-tuning on the 7B Instruct variant to directly teach the model to output ordered, typed event spans (start/end only) from raw clips without text queries.
\textbf{(3) Evaluation}: We evaluate the detector in two phases. First, we check its performance on a small curated subset from the source corpus, ensuring tight boundaries and full recovery of relevant GUI actions. Second, we apply the model to unseen web tutorials and conduct blinded manual review to assess its robustness and real-world usability. Details are provided in Appendix~\ref{app:denseeventdetection}.
% We assess the detector on two fronts. First, a small disjoint curation from the source corpus is used to check in-domain quality; qualitatively, the model produces tight boundaries and recovers the full set of salient GUI actions per clip. Second, we apply it to unseen web tutorials and conduct blinded manual review to verify robustness and usability in the wild. Implementation specifics are provided in Appendix~\ref{app:denseeventdetection}.

\subsubsection{Action Parameterization}\label{sec:method:actionidentification}
\textbf{Task.}
Given a detected action segment $v_k=v[t_k^{\mathrm{s}}:t_k^{\mathrm{e}}]$ with type $a_k\in\mathcal{A}$, predict the action content (parameters) $\pi_k$:
\[
h_\phi(v_k)\ \rightarrow\ (\hat{a}_k,\pi_k).
\]
For example, a click segment yields $h_\phi(v_k)\!\rightarrow\!(\text{click},(x,y))$, while a typing segment yields $h_\phi(v_k)\!\rightarrow\!(\text{type},\langle content \rangle)$.

% \textbf{Foundational capabilities.}To learn $h_\phi$ effectively, the recognizer must exhibit three core skills: \textbf{(1) cursor-centric tracking}—continuously follow the cursor across frames and resolve which UI element it engages with; \textbf{(2) screen-state change cognition}—detect and attribute visual transitions to specific actions (e.g., typed text appearing in an input field, content shifting due to a scroll); and \textbf{(3) fine-grained grounding}—precisely localize the target object (button, handle, text box, scrollable region) and align it with the cursor interaction to recover accurate parameters $\pi_k$.

\textbf{Approach.}
We build a recognizer $h_\phi$ that, for each detected segment $v_k$, predicts both the action type and its parameters $(\hat{a}_k,\pi_k)$. 
\textbf{(1) Training data}: We start from the OpenCUA raw demonstration logs, which pair screen-capture video with timestamped mouse and keyboard events. Each event is converted into type-specific parameter labels and temporally aligned to its clip, yielding prompt-free supervision that captures the exact content of the interaction. 
\textbf{(2) Model training}: Using Qwen2.5-VL (7B Instruct) as the base, we perform full-parameter supervised fine-tuning so the model maps $v_k$ directly to $(\hat{a}_k,\pi_k)$; when available, we optionally condition on the detector’s $a_k$ to stabilize type predictions. 
\textbf{(3) Evaluation}: Because ground-truth object boxes are unavailable, we evaluate only on unseen web tutorials via blinded manual review, assessing whether the predicted action type and parameters are correct and practically actionable.
Details are provided in Appendix~\ref{app:actionidentification}.

\subsubsection{Inner Monologue Generation}
Dense event detection and action identification recover what happened on screen but omit the stepwise rationale. We therefore generate a brief inner monologue $r_k$ before each action to make explicit the intent, the local plan, and the expected state change (e.g., “type query into the search box to reveal results,” “scroll to bring the ‘Settings’ button into view”). Explicit rationales provide structured supervision for planning and credit assignment, tie cursor–target grounding to goals and affordances, and improve robustness on long-horizon tasks via better error detection and recovery. Recent GUI-agent work that injects step-level “thoughts” or System-2 reasoning reports notable gains in perception, grounding, and task execution, motivating our inclusion of $r_k$ in ReAct-style trajectories \citep{xu2025aguvisunifiedpurevision,qin2025uitarspioneeringautomatedgui,wang2025opencuaopenfoundationscomputeruse}.

% \textbf{Approach.}
We cast inner-monologue generation as conditional paraphrasing with GPT-5 Medium. For each step $k$, we build a structured prompt that includes: (i) the detected action type $a_k$; (ii) its parameters $\pi_k$ (e.g., typed text, cursor coordinates); (iii) the screen state immediately before and after the action (keyframes or thumbnails); and (iv) short ASR transcripts spanning a 1-minute window before the action, the during span $[t_k^{\mathrm{s}},t_k^{\mathrm{e}}]$, and a 1-minute window after. Conditioned on these inputs, the model outputs a concise rationale $r_k$ that states the intent, the local plan, and the expected state change (grounded to visible UI). Additional prompt templates and representative inner-monologue examples are provided in Appendix~\ref{app:innermonologuegeneration}.

% \subsubsection{VideoAgentTrek Toolkit}
% \label{sec:method:toolkit}

% We release two core components as an open-source toolkit for video-based GUI understanding:

% \noindent\textbf{\screenfilter{}} tool enables the extraction of GUI segments from any video, efficiently removing non-interactive content with minimal computational overhead. Pre-trained models and inference code are provided for filtering GUI content.

% \noindent\textbf{\videoaction{}} offers a complete pipeline for trajectory extraction from GUI videos, including:
% (i) event detection for clicks, drags, scrolls, and typing;
% (ii) parameter inference for coordinates, directions, and text;
% (iii) action sequence generation in multiple output formats.
% The modular design allows using individual components or the full pipeline.

% Both tools are designed for standalone use or integration into larger systems.
% We provide comprehensive documentation, pretrained weights, and example scripts demonstrating common use cases.
% The toolkit enables researchers to extract GUI supervision from their own video sources without reimplementing our methods.

\subsection{Computer Use Model Pretraining}
\label{sec:method:building_computer_use_agent}

We demonstrate \textbf{\methodname}'s effectiveness by training an end-to-end computer-use agent with our video-driven data and a high-quality supervised finetuning set. On this strong finetuning basis, \textbf{\methodname} improves performance on online and offline agent evaluations.

\subsubsection{Agentic Data Collection}

\noindent\textbf{VideoAgentTrek Data.}
We apply \videoaction{} to the collected tutorial videos and convert them into agentic supervision. For each processed clip, we (i) run dense event detection to obtain typed, tightly bounded segments, (ii) infer action parameters with the action-identification recognizer, and (iii) generate a brief inner monologue for intent and expected state change. We then assemble the resulting steps $(I_k, r_k, a_k, \pi_k)$ into trajectories and serialize them for downstream training. In total, we processed 39{,}000 videos; each video produces on average 39 steps, yielding approximately 1.52 million ReAct steps overall, and about 26 billion training tokens. Detailed data statistics and examples will be provided in the Appendix~\ref{app:videoagenttrekdataanalysis}.

\noindent\textbf{Human demonstrations Data} We sample human-annotated trajectories from OpenCUA \citep{wang2025opencuaopenfoundationscomputeruse} and AGUVIS \citep{xu2025aguvisunifiedpurevision}, harmonizing formats and labels into a single schema. The corpus spans Windows, macOS, and Android, contributing about 8B tokens to training. 

\noindent\textbf{GUI Grounding Data.} We include a focused subset of GUI grounding pairs from the OSWorld-G dataset \citep{xie2025scalingcomputerusegroundinguser} to strengthen pointer–target alignment and layout-aware perception. This contributes roughly 1B tokens to training.

\subsubsection{Training strategy.}
Automatically mined trajectories, while large-scale, inevitably contain residual noise. Motivated by prior findings that decoupling perception/grounding from policy learning improves robustness \citep{xu2025aguvisunifiedpurevision,wang2025opencuaopenfoundationscomputeruse}, we adopt a two-stage schedule that first stabilizes grounding on broad but imperfect supervision and then consolidates policy on a clean subset.

\noindent\textbf{Foundation} Qwen2.5-VL-7B~\citep{bai2025qwen25vltechnicalreport} is a general vision-language model with superior vision understanding capability, but it is not sufficiently pretrained on computer-use tasks with an end-to-end success rate of 4.5\% on OSWorld~\citep{xie2024osworldbenchmarkingmultimodalagents}, which makes it a proper starting point (\textbf{base}) for evaluating the data generated by \methodname.

\noindent\textbf{Stage 1 training.}
We train for one epoch over \textbf{26B} tokens drawn from the VideoAgentTrek trajectories, augmented with a small number of GUI grounding pairs. Trajectories are formatted as interleaved vision–text sequences: frames (or frame-equivalent images) appear inline with the stepwise textual outputs, preserving temporal order across the entire clip. Loss is masked to the textual portions only; images are conditioning context and are not predicted. Please refer to Appendix~\ref{app:trainingstrategy} for representative formatting examples and complete training configurations including hardware, batch sizes, and optimization details.

\noindent\textbf{Stage 2 training.}
We continue training for \textbf{8B} tokens on a curated set of clean, human-annotated trajectories. Here we reformat the data into a chat template with user prompts and assistant responses that describe or execute the next action. We apply standard supervised finetuning with loss computed only on the assistant turns, leaving user turns as pure conditioning. Representative formatting examples and training details are provided in the Appendix~\ref{app:trainingstrategy}.

\section{Experiments}\label{sec:experiments}
\vspace{-5pt}
\subsection{Computer Use Agent Performance}
\label{sec:experiments:demonstrationofvideoagenttrek}
% We demonstrate the effectiveness of \textbf{\methodname} by training an end-to-end computer-use agent with our video-driven data pipeline and a high-quality supervised finetuning set. On this strong finetuning basis, \textbf{\methodname} further improves performance on both online and offline agent evaluations.
% \subsubsection{Experiment setup}

% \noindent\textbf{Model.} Qwen2.5-VL-7B~\citep{bai2025qwen25vltechnicalreport} is a general vision-language model with superior vision understanding capability, but it is not sufficiently pretrained on computer-use tasks with an end-to-end success rate of 4.5\% on OSWorld~\citep{xie2024osworldbenchmarkingmultimodalagents}, which makes it a proper starting point (\textbf{base}) for evaluating the data generated by \methodname.

\subsubsection{Experiment Setup.}

We evaluate the performance of our model on two computer-use agent benchmarks: OSWorld-Verified~\citep{osworld_verified,xie2024osworldbenchmarkingmultimodalagents} for online settings and AgentNetBench~\citep{wang2025opencuaopenfoundationscomputeruse} for offline settings. Further protocol, metrics, and computational details are provided in Appendix~\ref{app:computeruseagentevaluation}.

% We use two complementary benchmarks, OSWorld-Verified \citep{osworld_verified,xie2024osworldbenchmarkingmultimodalagents} and AgentNetBench \citep{wang2025opencuaopenfoundationscomputeruse}, to comprehensively evaluate the performance of our CUA under different computer systems and both online and offline settings. 
\begin{enumerate}[leftmargin=*]
    \item \textbf{OSWorld-Verified.} OSWorld~\citep{xie2024osworldbenchmarkingmultimodalagents} is an online computer-use agent evaluation benchmark that includes 369 human-crafted Ubuntu desktop tasks. OSWorld-Verified~\citep{osworld_verified} is a more stable version, with updated evaluation scripts, environments, and clarified instructions, designed to measure CUA's task-solving capabilities in dynamic, real-world environments.
    \item \textbf{AgentNetBench.} AgentNetBench is an offline benchmark is based on 100 representative tasks from the AgentNet dataset, covering a wide range of applications and websites on Windows and macOS. The tasks are manually refined and offer multiple valid action options for each step to reflect the variety of correct interactions. 
\end{enumerate}

\subsubsection{Main Results.}
\begin{figure}[t]
    % \vspace{-7mm}
  \centering
  \includegraphics[width=0.8\linewidth]{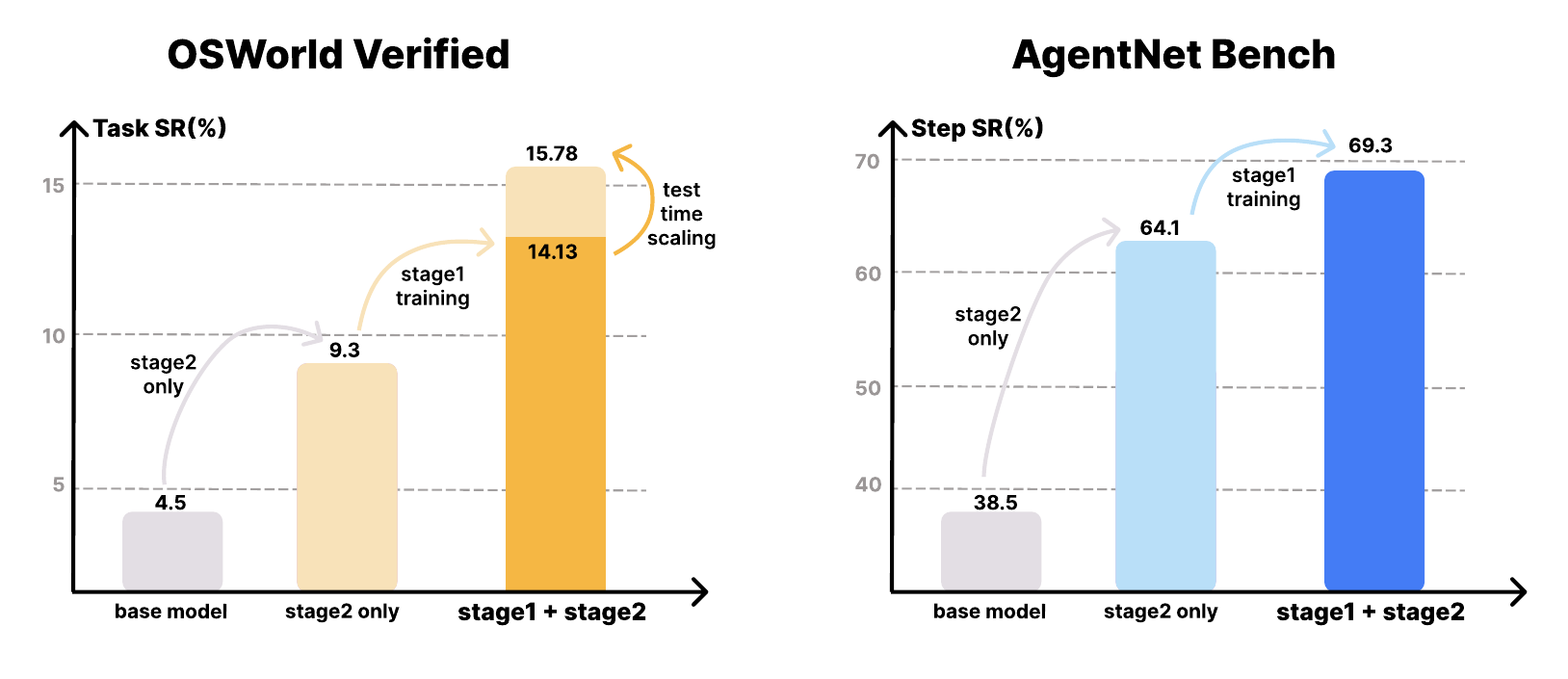}
\caption{Experimental Results on OSWorld-Verified~\citep{osworld_verified} and AgentNetBench~\citep{wang2025opencuaopenfoundationscomputeruse}. VideoAgentTrek demonstrates significant improvements over baseline models, with test-time scaling providing additional performance gains}
  \label{fig:cuaevaluationresults}
  % \vspace{-7mm}
\end{figure}

% \vspace{-1em}

% Figure~\ref{fig:cuaevaluationresults} compares computer-use agent performance across two benchmarks: task success rates on OSWorld-Verified and step success rates on AgentNetBench. Several key observations are as follows:

% \begin{enumerate}[leftmargin=1.4em]

\paragraph{Video pretraining enhances performance on offline benchmarks.}
    On AgentNetBench, incorporating VideoAgentTrek pretraining achieves a step success rate of \textbf{69.3\%}, representing a 5.2 percentage point improvement over the SFT-only baseline (64.1\%) and a substantial 30.8 percentage point gain over the base model (38.5\%). This consistent improvement demonstrates that video pretraining effectively transfers knowledge to structured offline evaluation scenarios.

\paragraph{Video pretraining delivers greater improvements on online benchmarks.}  
      On OSWorld-Verified, our complete approach achieves a task success rate of 14.13\%, demonstrating a 4.83 percentage point improvement (+52\% relative) over SFT-only training (9.3\%) and more than tripling the performance of the base model (4.5\%). % The larger improvement margin in the online setting—where agents must handle real-time interactions and environmental uncertainties—indicates that video pretraining enhances robustness for dynamic interactive scenarios that more closely resemble real-world deployment conditions.

\paragraph{Video pretraining enables effective test-time scaling for computer-use agents.}  
      As shown in Figure~\ref{fig:cuaevaluationresults}, the performance of model trained with stage 1 and stage 2 improved from 14.13\% to 15.78 \% when the action step budget increases from 20 to 50 steps on OSWorld-Verified. This 1.65 percentage point improvement demonstrates the model's ability to effectively utilize additional exploration opportunities. This test-time scaling benefit emerges specifically from our video pretraining: models trained on longer video trajectories learn to effectively utilize extended planning horizons, while the SFT-only baseline shows no improvement with additional steps (see Section~\ref{sec:analysis:effectivenessofdatascaling}).

% \input{tables/cua_results}
% On OSWorld-Verified, the base model achieves a task success rate (SR) of 4.5\%; stage 2 (SFT-only) increases this to 9.3\%.
% Introducing VideoAgentTrek stage 1 raises the SR to 14.13\%.
% When we apply test-time scaling by expanding the action step budget from 20 to 50 steps, allowing the agent more opportunities to explore, performance further improves to \textbf{15.78\%}.
% This test-time scaling benefit emerges specifically from our video pretraining: models trained on longer video trajectories learn to effectively utilize extended planning horizons, while the SFT-only baseline shows no improvement with additional steps (see Section~\ref{sec:analysis:effectivenessofdatascaling}).

% On OSWorld-Verified, the base model attains 4.5\% task SR; SFT-only (stage~2) roughly doubles this to 9.3\%.
% Adding our VideoAgentTrek stage~1 yields 14.13\% (+4.83 points over stage~2, +52\% relative), and simple test-time scaling further lifts performance to \textbf{15.78\%} (+1.65).
% On AgentNetBench (offline), step SR rises from 38.5\% (base) to 64.1\% with stage~2, and to \textbf{69.3\%} with stage~1\,+\,stage~2 (+5.2 points over stage~2).
% Overall, VideoAgentTrek pretraining consistently improves over a strong finetuning baseline, with especially large gains in the online setting, indicating better robustness and long-horizon competence.

\subsection{\videoaction{} Performance}
\subsubsection{Action Event Detection}\label{sec:experiments:denseeventdetection}
We assess \videoaction{} with a two-part protocol: a held-out, annotated test set and an in-the-wild manual validation:
% \begin{enumerate}[leftmargin=*]
%     \item \textbf{Held-out test set.} We hold out 23 hours of screen-capture videos with 20{,}282 annotated GUI events. Each event is a tuple \((\text{type}, t^{\mathrm{s}}, t^{\mathrm{e}})\). A prediction counts as a \emph{hit} iff its type matches and its interval has any temporal overlap with a ground-truth event; unmatched predictions are false positives and unmatched ground truths are false negatives. We report per-type Precision/Recall/F1 and micro/macro aggregates.
%     \item \textbf{Manual validation (in-the-wild).} On 10 unseen YouTube tutorials, we apply the same overlap criterion and estimate recovery rates by human review to assess robustness outside the curated set.
% \end{enumerate}
\paragraph{Held-out test set.} We hold out 23 hours of screen-capture videos with 20{,}282 annotated GUI events. Each event is a tuple \((\text{type}, t^{\mathrm{s}}, t^{\mathrm{e}})\). A prediction counts as a \emph{hit} iff its type matches and its interval has any temporal overlap with a ground-truth event; unmatched predictions are false positives and unmatched ground truths are false negatives. We report per-type Precision/Recall/F1 and micro/macro aggregates.
\paragraph{Manual validation (in-the-wild).} On 10 unseen YouTube tutorials, we apply the same overlap criterion and estimate recovery rates by human review to assess robustness outside the curated set.

\begin{table}[t]
  % \vspace{-5mm}
  \centering
  \setlength{\tabcolsep}{4pt}
  % --- 左表 ---
  \begin{minipage}[t]{0.5\linewidth}
    \centering
    \small
    \resizebox{\linewidth}{!}{%
    \begin{tabular}{lrrrr>{\columncolor{LightBlue}}r}
    \toprule
    \textbf{Action} & \textbf{Preds} & \textbf{GT} & \textbf{Precision} & \textbf{Recall} & \textbf{F1} \\
    \midrule
    Click  & 12{,}222 & 14{,}247 & 0.88 & 0.76 & 0.82 \\
    Drag   & 971      & 1{,}462  & 0.78 & 0.52 & 0.62 \\
    Press  & 177      & 842      & 0.40 & 0.08 & 0.14 \\
    Scroll & 1{,}448  & 1{,}691  & 0.93 & 0.80 & 0.86 \\
    Type   & 1{,}480  & 2{,}040  & 0.89 & 0.64 & 0.75 \\
    \midrule
    \textbf{Total} & \textbf{17{,}298} & \textbf{20{,}282} & \textbf{0.88} & \textbf{0.70} & \textbf{0.78} \\
    \bottomrule
    \end{tabular}%
    }
    \caption{Action-event detector evaluation: held-out test-set results by action type.}
    \label{tab:denseeventdetectorevaluation}
  \end{minipage}
  \hfill
  % --- 右表 ---
  \begin{minipage}[t]{0.45\linewidth}
    \centering
    \small
    % \resizebox{\linewidth}{!}{%
    \begin{tabular}{l r r}
    \toprule
    \textbf{Action type} & \textbf{Samples} & \textbf{Accuracy} \\
    \midrule
    Click  & 324 & 0.713 \\
    Drag   & 22  & 0.366 \\
    Press  & 47  & 0.362 \\
    Scroll & 34  & 0.735 \\
    Type   & 73  & 0.671 \\
    \midrule
    \textbf{Overall} & \textbf{500} & \textbf{0.658} \\
    \bottomrule
    \end{tabular}
    % }
    \caption{Action parameterization evaluation: manual in-the-wild assessment.}
    \label{tab:action-id-manual}
  \end{minipage}
  % \vspace{-5mm}
\end{table}

\noindent\textbf{Results.}
As the results shown in Table~\ref{tab:denseeventdetectorevaluation}, Overall precision is high (0.88) with solid recall (0.71). Pointer-centric actions (click, scroll) are reliably localized; keystroke-only actions show lower recall/precision due to subtle visual evidence. In the manual study, the detector recovers \(\sim\)70\% of actions under the same criterion, consistent with in-house results.

\subsubsection{Action Identification}\label{sec:experiments:actionidentification}

Evaluating action identification automatically is difficult because target-element boxes are unavailable. We therefore apply the identifier to in-the-wild videos and perform a blinded manual assessment. An action is judged \emph{proper} if, when executed, it would plausibly produce the observed on-screen transition (for example, the clicked control changes state, typed text appears in the focused field, or the page scrolls). We evaluate 500 predictions sampled across action types.

\noindent\textbf{Results.}
Annotators review pre/post frames and verify whether predicted parameters explain the observed changes, with disagreements resolved through second-pass review. Performance varies by action type: pointer-based actions (click, scroll) achieve highest accuracy, typing shows moderate accuracy despite OCR noise, while drag/press actions struggle with subtle visual cues. Despite these challenges, the predicted parameters are accurate enough for trajectory construction and downstream training; detailed counts and validation rates appear in Table~\ref{tab:action-id-manual}.

\section{Analysis}\label{sec:analysis}

\subsection{Effectiveness of Data Scaling}\label{sec:analysis:effectivenessofdatascaling}
% \noindent\textbf{Effect of Stage-1 scale on downstream performance.}

\begin{wrapfigure}{r}{0.35\textwidth} % narrower width; more wrapped lines
\vspace{-55pt}
\centering
\includegraphics[width=\linewidth]{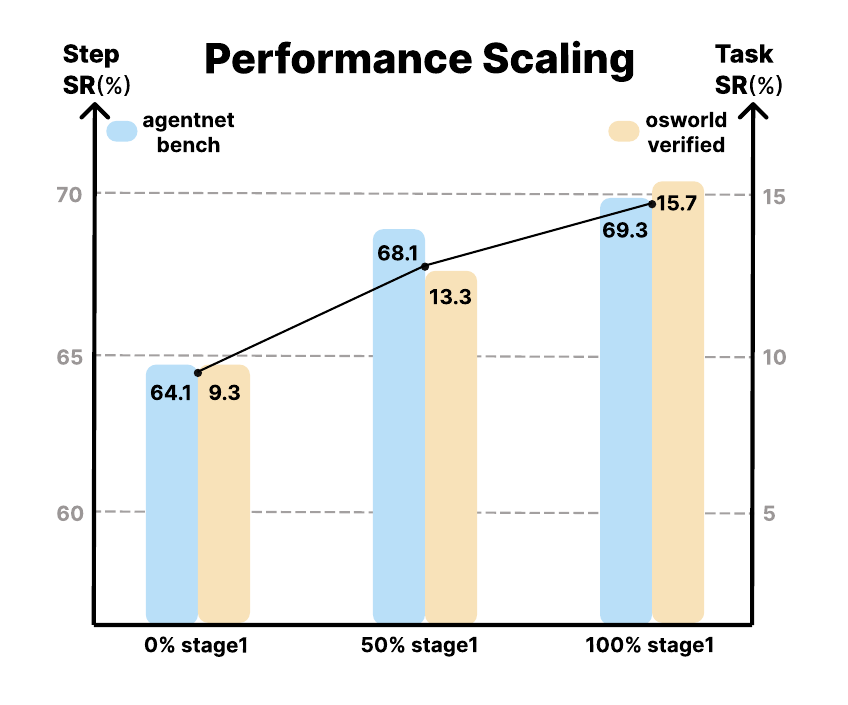}
\caption{Performance Scaling}
\label{fig:stage1_scaling}
\vspace{-2em}
\end{wrapfigure}

To assess the impact of Stage-1 data scale, we train models using 0\%, 50\%, and 100\% of the video tokens, then apply identical Stage-2 SFT to each variant. With increasing tokens, performance scales consistently across both benchmarks. AgentNetBench step success rates increase from 64.1\% to 68.1\% and 69.3\%, while OSWorld-Verified task SR@50 grow from 9.3\% to 13.3\% and 15.7\% (Figure~\ref{fig:stage1_scaling}).
These findings establish a clear relationship between pretraining data size and computer-use agent performance, demonstrating the benefits of scaling video pretraining data.

\subsection{Improving Long Horizon planning}
\methodname \space provides substantially longer trajectories than previous CUA corpora. As illustrated in Figure~\ref{fig:videoagenttrekdatadistributionofstepnumber}, 42.1\% of trajectories exceed 20 steps, while 14.5\% contain 50 or more steps, yielding an average trajectory length of 39.25 steps. Cross-dataset comparisons (Table~\ref{tab:dataset_step_comparison}) reveal that this average substantially exceeds those of established benchmarks, demonstrating that \methodname corpus emphasizes supervision of complex, multi-step workflows rather than brief, single-interaction sequences.

The benefits of long-horizon supervision become evident when evaluating planning capabilities under varying step budgets.
On OSWorld-Verified, we observe a striking difference in how models respond to increased action budgets.
The Stage2-only model shows no performance improvement when the step budget expands from 20 to 50 steps, remaining flat at 9.3\% task success, indicating it cannot effectively plan beyond its training horizon or recover from early mistakes.
In contrast, after Stage-1 pretraining on VideoAgentTrek's long video trajectories, the agent demonstrates true test-time scaling: task success rate increases from 14.13\% at 20 steps to 15.78\% at 50 steps, a +1.65 point absolute improvement (+11.7\% relative gain, Figure~\ref{fig:cuaevaluationresults}).
This differential reveals that exposure to extended video demonstrations during pretraining teaches the model to decompose complex tasks into subgoals, persist through intermediate failures, and leverage additional computational budget for exploration and error correction, capabilities that supervised fine-tuning on shorter trajectories fails to instill.

\section{Related Work}\label{sec:related}

\noindent\textbf{Generating agent trajectories.}
Computer-use trajectories have been obtained through human annotation, programmatic synthesis in instrumented environments, and web-scale mining of public resources. Human annotation, often aided by instrumentation to log pointer coordinates and keystrokes, yields precise labels but is costly and narrow in coverage \citep{qin2025uitarspioneeringautomatedgui,wang2025uitars2technicalreportadvancing,wang2025opencuaopenfoundationscomputeruse}. Programmatic synthesis inside headless browsers or scripted desktop flows can generate large volumes with exact parameters, yet coverage is constrained by simulator APIs and may diverge from real-world UI variability \citep{su2025learnbyinteractdatacentricframeworkselfadaptive,sun2025osgenesisautomatingguiagent}. Web-scale mining taps tutorials, RPA logs, and screen recordings to obtain diverse trajectories, but typically lacks precise temporal boundaries or action parameters \citep{xu2025agenttrekagenttrajectorysynthesis,jang2025scalablevideotodatasetgenerationcrossplatform}.

\noindent\textbf{Precise Event Grounding in Video.}
Temporal grounding approaches such as temporal action localization, moment retrieval, keyframe detection, and dense video captioning seek to determine when events take place and provide corresponding descriptions \citep{lin2019bmnboundarymatchingnetworktemporal,zhuang2025groundingmdgroundedvideolanguagepretraining,wasim2024videogroundingdinoopenvocabularyspatiotemporalvideo}. Meanwhile, recent multimodal systems (e.g., Qwen2.5-VL~\citep{bai2025qwen25vltechnicalreport}, Gemini 2.5 Pro~\citep{comanici2025gemini25pushingfrontier}) have advanced the field by enabling more detailed spatiotemporal understanding and long-horizon video reasoning. Nonetheless, most general-purpose grounding frameworks focus primarily on semantic interpretation, rather than achieving the millisecond-level precision and parameter extraction required to faithfully reconstruct GUI interactions.

\noindent\textbf{Learning from unlabeled video to act in environments.}
VPT demonstrated that large-scale \emph{unlabeled} videos can be converted into effective training signals (e.g., via inverse-dynamics auto-labeling followed by behavior cloning), substantially improving an agent's ability to act \citep{baker2022videopretrainingvptlearning}.Building on this idea, subsequent work leverages internet-scale human videos to distill human policy priors that transfer to interactive environments, including learning action-centric latent spaces without action labels \citep{ye2025latentactionpretrainingvideos} and scaling to humanoid control \citep{mao2024learningmassivehumanvideos}.

\section{Conclusion}\label{sec:conclusion}
We presented VideoAgentTrek, a scalable pipeline that transforms publicly available screen recordings into structured supervision for computer-use agents without manual annotation. By developing an inverse dynamics module that accurately detects GUI events and extracts action parameters from raw video, we demonstrate that the implicit supervision in tutorial videos can be effectively harvested at scale. Our experiments on 39,000 YouTube videos yielded 1.52 million interaction steps, enabling continued pretraining that improved task success rates by 70\% on OSWorld-Verified (9.3\% to 15.8\%) and increased step accuracy on AgentNetBench from 64.1\% to 69.3\%. These results establish that unlabeled internet videos, when processed through learned inverse dynamics, provide a viable and cost-effective alternative to expensive manual annotation for training robust computer-use agents. The open-source release of our ScreenFilter and Video2Action tools enables the community to leverage this abundant resource for advancing GUI automation research.

\clearpage
\bibliography{colm2024_conference}

\begin{thebibliography}{26}
\providecommand{\natexlab}[1]{#1}
\providecommand{\url}[1]{\texttt{#1}}
\expandafter\ifx\csname urlstyle\endcsname\relax
  \providecommand{\doi}[1]{doi: #1}\else
  \providecommand{\doi}{doi: \begingroup \urlstyle{rm}\Url}\fi

\bibitem[Bai et~al.(2025)Bai, Chen, Liu, Wang, Ge, Song, Dang, Wang, Wang, Tang, Zhong, Zhu, Yang, Li, Wan, Wang, Ding, Fu, Xu, Ye, Zhang, Xie, Cheng, Zhang, Yang, Xu, and Lin]{bai2025qwen25vltechnicalreport}
Shuai Bai, Keqin Chen, Xuejing Liu, Jialin Wang, Wenbin Ge, Sibo Song, Kai Dang, Peng Wang, Shijie Wang, Jun Tang, Humen Zhong, Yuanzhi Zhu, Mingkun Yang, Zhaohai Li, Jianqiang Wan, Pengfei Wang, Wei Ding, Zheren Fu, Yiheng Xu, Jiabo Ye, Xi~Zhang, Tianbao Xie, Zesen Cheng, Hang Zhang, Zhibo Yang, Haiyang Xu, and Junyang Lin.
\newblock Qwen2.5-vl technical report, 2025.
\newblock URL \url{https://arxiv.org/abs/2502.13923}.

\bibitem[Baker et~al.(2022)Baker, Akkaya, Zhokhov, Huizinga, Tang, Ecoffet, Houghton, Sampedro, and Clune]{baker2022videopretrainingvptlearning}
Bowen Baker, Ilge Akkaya, Peter Zhokhov, Joost Huizinga, Jie Tang, Adrien Ecoffet, Brandon Houghton, Raul Sampedro, and Jeff Clune.
\newblock Video pretraining (vpt): Learning to act by watching unlabeled online videos, 2022.
\newblock URL \url{https://arxiv.org/abs/2206.11795}.

\bibitem[Chen et~al.(2025)Chen, Cui, Hu, Qin, Fang, Zhao, Wang, Liu, Chen, Huo, Yao, Lin, Liu, and Sun]{chen2025guicoursegeneralvisionlanguage}
Wentong Chen, Junbo Cui, Jinyi Hu, Yujia Qin, Junjie Fang, Yue Zhao, Chongyi Wang, Jun Liu, Guirong Chen, Yupeng Huo, Yuan Yao, Yankai Lin, Zhiyuan Liu, and Maosong Sun.
\newblock Guicourse: From general vision language models to versatile gui agents, 2025.
\newblock URL \url{https://arxiv.org/abs/2406.11317}.

\bibitem[Comanici et~al.(2025)Comanici, Bieber, Schaekermann, and Pasupat.etc]{comanici2025gemini25pushingfrontier}
Gheorghe Comanici, Eric Bieber, Mike Schaekermann, and Ice Pasupat.etc.
\newblock Gemini 2.5: Pushing the frontier with advanced reasoning, multimodality, long context, and next generation agentic capabilities, 2025.
\newblock URL \url{https://arxiv.org/abs/2507.06261}.

\bibitem[Deng et~al.(2023)Deng, Gu, Zheng, Chen, Stevens, Wang, Sun, and Su]{deng2023mind2webgeneralistagentweb}
Xiang Deng, Yu~Gu, Boyuan Zheng, Shijie Chen, Samuel Stevens, Boshi Wang, Huan Sun, and Yu~Su.
\newblock Mind2web: Towards a generalist agent for the web, 2023.
\newblock URL \url{https://arxiv.org/abs/2306.06070}.

\bibitem[Jang et~al.(2025)Jang, Song, Sohn, Logeswaran, Luo, Kim, Bae, and Lee]{jang2025scalablevideotodatasetgenerationcrossplatform}
Yunseok Jang, Yeda Song, Sungryull Sohn, Lajanugen Logeswaran, Tiange Luo, Dong-Ki Kim, Kyunghoon Bae, and Honglak Lee.
\newblock Scalable video-to-dataset generation for cross-platform mobile agents, 2025.
\newblock URL \url{https://arxiv.org/abs/2505.12632}.

\bibitem[Li et~al.(2024)Li, Bishop, Li, Rawles, Campbell-Ajala, Tyamagundlu, and Riva]{li2024effectsdatascaleui}
Wei Li, William Bishop, Alice Li, Chris Rawles, Folawiyo Campbell-Ajala, Divya Tyamagundlu, and Oriana Riva.
\newblock On the effects of data scale on ui control agents, 2024.
\newblock URL \url{https://arxiv.org/abs/2406.03679}.

\bibitem[Lin et~al.(2019)Lin, Liu, Li, Ding, and Wen]{lin2019bmnboundarymatchingnetworktemporal}
Tianwei Lin, Xiao Liu, Xin Li, Errui Ding, and Shilei Wen.
\newblock Bmn: Boundary-matching network for temporal action proposal generation, 2019.
\newblock URL \url{https://arxiv.org/abs/1907.09702}.

\bibitem[Mao et~al.(2024)Mao, Zhao, Song, Shi, Ye, Zhang, Geng, Malik, Guizilini, and Wang]{mao2024learningmassivehumanvideos}
Jiageng Mao, Siheng Zhao, Siqi Song, Tianheng Shi, Junjie Ye, Mingtong Zhang, Haoran Geng, Jitendra Malik, Vitor Guizilini, and Yue Wang.
\newblock Learning from massive human videos for universal humanoid pose control, 2024.
\newblock URL \url{https://arxiv.org/abs/2412.14172}.

\bibitem[Nguyen-Tuong \& Peters(2010)Nguyen-Tuong and Peters]{nguyen2010using}
Duy Nguyen-Tuong and Jan Peters.
\newblock Using model knowledge for learning inverse dynamics.
\newblock In \emph{2010 IEEE international conference on robotics and automation}, pp.\  2677--2682. IEEE, 2010.

\bibitem[Qin et~al.(2025)Qin, Ye, Fang, Wang, Liang, Tian, Zhang, Li, Li, Huang, Zhong, Li, Yang, Miao, Lin, Liu, Jiang, Ma, Li, Xiao, Cai, Li, Zheng, Jin, Li, Zhou, Wang, Chen, Li, Yang, Liu, Lin, Peng, Liu, and Shi]{qin2025uitarspioneeringautomatedgui}
Yujia Qin, Yining Ye, Junjie Fang, Haoming Wang, Shihao Liang, Shizuo Tian, Junda Zhang, Jiahao Li, Yunxin Li, Shijue Huang, Wanjun Zhong, Kuanye Li, Jiale Yang, Yu~Miao, Woyu Lin, Longxiang Liu, Xu~Jiang, Qianli Ma, Jingyu Li, Xiaojun Xiao, Kai Cai, Chuang Li, Yaowei Zheng, Chaolin Jin, Chen Li, Xiao Zhou, Minchao Wang, Haoli Chen, Zhaojian Li, Haihua Yang, Haifeng Liu, Feng Lin, Tao Peng, Xin Liu, and Guang Shi.
\newblock Ui-tars: Pioneering automated gui interaction with native agents, 2025.
\newblock URL \url{https://arxiv.org/abs/2501.12326}.

\bibitem[Reis et~al.(2024)Reis, Kupec, Hong, and Daoudi]{reis2024realtimeflyingobjectdetection}
Dillon Reis, Jordan Kupec, Jacqueline Hong, and Ahmad Daoudi.
\newblock Real-time flying object detection with yolov8, 2024.
\newblock URL \url{https://arxiv.org/abs/2305.09972}.

\bibitem[Su et~al.(2025)Su, Sun, Yoon, Yin, Yu, and Arık]{su2025learnbyinteractdatacentricframeworkselfadaptive}
Hongjin Su, Ruoxi Sun, Jinsung Yoon, Pengcheng Yin, Tao Yu, and Sercan~Ö. Arık.
\newblock Learn-by-interact: A data-centric framework for self-adaptive agents in realistic environments, 2025.
\newblock URL \url{https://arxiv.org/abs/2501.10893}.

\bibitem[Sun et~al.(2025)Sun, Cheng, Ding, Jin, Wang, Xu, Wu, Jia, Chen, Liu, Kao, Li, He, Qiao, and Wu]{sun2025osgenesisautomatingguiagent}
Qiushi Sun, Kanzhi Cheng, Zichen Ding, Chuanyang Jin, Yian Wang, Fangzhi Xu, Zhenyu Wu, Chengyou Jia, Liheng Chen, Zhoumianze Liu, Ben Kao, Guohao Li, Junxian He, Yu~Qiao, and Zhiyong Wu.
\newblock Os-genesis: Automating gui agent trajectory construction via reverse task synthesis, 2025.
\newblock URL \url{https://arxiv.org/abs/2412.19723}.

\bibitem[Team et~al.(2025)Team, Du, Yin, Xing, Qu, Wang, Chen, Zhang, Du, Wei, Wang, Zhang, Du, Wang, Yuan, Lu, Li, Sung, Wei, Lai, Zhu, Ding, Hu, Yang, Zhang, Wu, Yao, Lu, Wang, Gao, Zheng, Li, Su, Wang, Deng, Qiu, Xie, Wang, Liu, Yan, Ouyang, Chen, Sui, Yu, Dong, Dong, Xu, Cheng, Gu, Zhou, Liu, Cao, Yu, Song, Bai, Song, He, Huang, Xu, Yuan, Yao, Wu, Li, Zu, Zhou, Wang, Charles, Zhong, Li, Hu, Chen, Wang, Liu, Miao, Qin, Chen, Bao, Wang, Kang, Liu, Dong, Du, Wu, Wang, Yan, Zhou, Li, Jiang, Zhang, Yang, Huang, Huang, Zhao, Chen, and Lin]{kimiteam2025kimivltechnicalreport}
Kimi Team, Angang Du, Bohong Yin, Bowei Xing, Bowen Qu, Bowen Wang, Cheng Chen, Chenlin Zhang, Chenzhuang Du, Chu Wei, Congcong Wang, Dehao Zhang, Dikang Du, Dongliang Wang, Enming Yuan, Enzhe Lu, Fang Li, Flood Sung, Guangda Wei, Guokun Lai, Han Zhu, Hao Ding, Hao Hu, Hao Yang, Hao Zhang, Haoning Wu, Haotian Yao, Haoyu Lu, Heng Wang, Hongcheng Gao, Huabin Zheng, Jiaming Li, Jianlin Su, Jianzhou Wang, Jiaqi Deng, Jiezhong Qiu, Jin Xie, Jinhong Wang, Jingyuan Liu, Junjie Yan, Kun Ouyang, Liang Chen, Lin Sui, Longhui Yu, Mengfan Dong, Mengnan Dong, Nuo Xu, Pengyu Cheng, Qizheng Gu, Runjie Zhou, Shaowei Liu, Sihan Cao, Tao Yu, Tianhui Song, Tongtong Bai, Wei Song, Weiran He, Weixiao Huang, Weixin Xu, Xiaokun Yuan, Xingcheng Yao, Xingzhe Wu, Xinhao Li, Xinxing Zu, Xinyu Zhou, Xinyuan Wang, Y.~Charles, Yan Zhong, Yang Li, Yangyang Hu, Yanru Chen, Yejie Wang, Yibo Liu, Yibo Miao, Yidao Qin, Yimin Chen, Yiping Bao, Yiqin Wang, Yongsheng Kang, Yuanxin Liu, Yuhao Dong, Yulun Du, Yuxin Wu, Yuzhi Wang, Yuzi Yan, Zaida
  Zhou, Zhaowei Li, Zhejun Jiang, Zheng Zhang, Zhilin Yang, Zhiqi Huang, Zihao Huang, Zijia Zhao, Ziwei Chen, and Zongyu Lin.
\newblock Kimi-vl technical report, 2025.
\newblock URL \url{https://arxiv.org/abs/2504.07491}.

\bibitem[Wang et~al.(2025{\natexlab{a}})Wang, Zou, Song, Feng, Fang, Lu, Liu, ..., Zhao, and Shi]{wang2025uitars2technicalreportadvancing}
Haoming Wang, Haoyang Zou, Huatong Song, Jiazhan Feng, Junjie Fang, Junting Lu, Longxiang Liu, ..., Qinghao Zhao, and Guang Shi.
\newblock Ui-tars-2 technical report: Advancing gui agent with multi-turn reinforcement learning, 2025{\natexlab{a}}.
\newblock URL \url{https://arxiv.org/abs/2509.02544}.

\bibitem[Wang et~al.(2025{\natexlab{b}})Wang, Wang, Lu, Yang, Xie, Wang, Deng, Guo, Xu, Wu, Shen, Li, Li, Li, Chen, Zheng, Li, Lei, Cao, Fu, Shin, Shin, Hu, Wang, Chen, Ye, Zhang, Du, Hu, Chen, Zhou, Yao, Chen, Gu, Wang, Wang, Yang, Zhong, Sung, Charles, Yang, and Yu]{wang2025opencuaopenfoundationscomputeruse}
Xinyuan Wang, Bowen Wang, Dunjie Lu, Junlin Yang, Tianbao Xie, Junli Wang, Jiaqi Deng, Xiaole Guo, Yiheng Xu, Chen~Henry Wu, Zhennan Shen, Zhuokai Li, Ryan Li, Xiaochuan Li, Junda Chen, Boyuan Zheng, Peihang Li, Fangyu Lei, Ruisheng Cao, Yeqiao Fu, Dongchan Shin, Martin Shin, Jiarui Hu, Yuyan Wang, Jixuan Chen, Yuxiao Ye, Danyang Zhang, Dikang Du, Hao Hu, Huarong Chen, Zaida Zhou, Haotian Yao, Ziwei Chen, Qizheng Gu, Yipu Wang, Heng Wang, Diyi Yang, Victor Zhong, Flood Sung, Y.~Charles, Zhilin Yang, and Tao Yu.
\newblock Opencua: Open foundations for computer-use agents, 2025{\natexlab{b}}.
\newblock URL \url{https://arxiv.org/abs/2508.09123}.

\bibitem[Wasim et~al.(2024)Wasim, Naseer, Khan, Yang, and Khan]{wasim2024videogroundingdinoopenvocabularyspatiotemporalvideo}
Syed~Talal Wasim, Muzammal Naseer, Salman Khan, Ming-Hsuan Yang, and Fahad~Shahbaz Khan.
\newblock Video-groundingdino: Towards open-vocabulary spatio-temporal video grounding, 2024.
\newblock URL \url{https://arxiv.org/abs/2401.00901}.

\bibitem[Xie et~al.(2024)Xie, Zhang, Chen, Li, Zhao, Cao, Hua, Cheng, Shin, Lei, Liu, Xu, Zhou, Savarese, Xiong, Zhong, and Yu]{xie2024osworldbenchmarkingmultimodalagents}
Tianbao Xie, Danyang Zhang, Jixuan Chen, Xiaochuan Li, Siheng Zhao, Ruisheng Cao, Toh~Jing Hua, Zhoujun Cheng, Dongchan Shin, Fangyu Lei, Yitao Liu, Yiheng Xu, Shuyan Zhou, Silvio Savarese, Caiming Xiong, Victor Zhong, and Tao Yu.
\newblock Osworld: Benchmarking multimodal agents for open-ended tasks in real computer environments, 2024.
\newblock URL \url{https://arxiv.org/abs/2404.07972}.

\bibitem[Xie et~al.(2025{\natexlab{a}})Xie, Deng, Li, Yang, Wu, Chen, Hu, Wang, Xu, Wang, Xu, Wang, Sahoo, Yu, and Xiong]{xie2025scalingcomputerusegroundinguser}
Tianbao Xie, Jiaqi Deng, Xiaochuan Li, Junlin Yang, Haoyuan Wu, Jixuan Chen, Wenjing Hu, Xinyuan Wang, Yuhui Xu, Zekun Wang, Yiheng Xu, Junli Wang, Doyen Sahoo, Tao Yu, and Caiming Xiong.
\newblock Scaling computer-use grounding via user interface decomposition and synthesis, 2025{\natexlab{a}}.
\newblock URL \url{https://arxiv.org/abs/2505.13227}.

\bibitem[Xie et~al.(2025{\natexlab{b}})Xie, Yuan, Zhang, Xiong, Shen, Zhou, Wang, Chen, Deng, Chen, Wang, Wu, Chen, Wang, Lu, Hu, and Yu]{osworld_verified}
Tianbao Xie, Mengqi Yuan, Danyang Zhang, Xinzhuang Xiong, Zhennan Shen, Zilong Zhou, Xinyuan Wang, Yanxu Chen, Jiaqi Deng, Junda Chen, Bowen Wang, Haoyuan Wu, Jixuan Chen, Junli Wang, Dunjie Lu, Hao Hu, and Tao Yu.
\newblock Introducing osworld-verified.
\newblock \emph{xlang.ai}, July 2025{\natexlab{b}}.
\newblock URL \url{https://xlang.ai/blog/osworld-verified}.

\bibitem[Xu et~al.(2025{\natexlab{a}})Xu, Lu, Shen, Wang, Wang, Mao, Xiong, and Yu]{xu2025agenttrekagenttrajectorysynthesis}
Yiheng Xu, Dunjie Lu, Zhennan Shen, Junli Wang, Zekun Wang, Yuchen Mao, Caiming Xiong, and Tao Yu.
\newblock Agenttrek: Agent trajectory synthesis via guiding replay with web tutorials, 2025{\natexlab{a}}.
\newblock URL \url{https://arxiv.org/abs/2412.09605}.

\bibitem[Xu et~al.(2025{\natexlab{b}})Xu, Wang, Wang, Lu, Xie, Saha, Sahoo, Yu, and Xiong]{xu2025aguvisunifiedpurevision}
Yiheng Xu, Zekun Wang, Junli Wang, Dunjie Lu, Tianbao Xie, Amrita Saha, Doyen Sahoo, Tao Yu, and Caiming Xiong.
\newblock Aguvis: Unified pure vision agents for autonomous gui interaction, 2025{\natexlab{b}}.
\newblock URL \url{https://arxiv.org/abs/2412.04454}.

\bibitem[Yao et~al.(2023)Yao, Zhao, Yu, Du, Shafran, Narasimhan, and Cao]{react}
Shunyu Yao, Jeffrey Zhao, Dian Yu, Nan Du, Izhak Shafran, Karthik~R. Narasimhan, and Yuan Cao.
\newblock React: Synergizing reasoning and acting in language models.
\newblock In \emph{The Eleventh International Conference on Learning Representations, {ICLR} 2023, Kigali, Rwanda, May 1-5, 2023}. OpenReview.net, 2023.

\bibitem[Ye et~al.(2025)Ye, Jang, Jeon, Joo, Yang, Peng, Mandlekar, Tan, Chao, Lin, Liden, Lee, Gao, Zettlemoyer, Fox, and Seo]{ye2025latentactionpretrainingvideos}
Seonghyeon Ye, Joel Jang, Byeongguk Jeon, Sejune Joo, Jianwei Yang, Baolin Peng, Ajay Mandlekar, Reuben Tan, Yu-Wei Chao, Bill~Yuchen Lin, Lars Liden, Kimin Lee, Jianfeng Gao, Luke Zettlemoyer, Dieter Fox, and Minjoon Seo.
\newblock Latent action pretraining from videos, 2025.
\newblock URL \url{https://arxiv.org/abs/2410.11758}.

\bibitem[Zhuang et~al.(2025)Zhuang, Li, Li, Liu, Hong, Gao, Yang, and Zuo]{zhuang2025groundingmdgroundedvideolanguagepretraining}
Weijun Zhuang, Qizhang Li, Xin Li, Ming Liu, Xiaopeng Hong, Feng Gao, Fan Yang, and Wangmeng Zuo.
\newblock Grounding-md: Grounded video-language pre-training for open-world moment detection, 2025.
\newblock URL \url{https://arxiv.org/abs/2504.14553}.

\end{thebibliography}
\bibliographystyle{colm2024_conference}

\clearpage
\appendix

\section{YouTube Video Quality Standards}
\label{app:youtubevideoevaluation}

To ensure consistency and usability in selecting high-quality instructional videos from YouTube for research purposes, the following standards must be met:

\begin{enumerate}[label=\arabic*.]
    \item \textbf{Minimal Overlays.} If overlays, such as face cams or titles, are present, they must occupy no more than $\frac{1}{10}$ of the screen area to avoid obstructing the primary content.
    \item \textbf{Primary Focus on Screen Recording.} The video should predominantly feature clean screen recordings. Brief transitions to other scenes, such as PowerPoint slides or face capture, are permissible but should be limited to introductory or concluding segments.
    \item \textbf{Screen Recording Method.} The video must consist of direct screen recordings rather than footage captured by an external camera.
    \item \textbf{Language Requirement.} The video must be in English to facilitate monolingual captioning in subsequent processing steps.
    \item \textbf{Stable Visual Presentation.} Frequent zooming in or out should be avoided. The entire screen or application window must be visible for the majority of the video duration.
    \item \textbf{Caption Availability.} The video must include captions, indicated by the availability of the closed caption (CC) icon in the bottom right corner of the player. Captions may be auto-generated or manually annotated.
    \item \textbf{Orientation.} The video must be recorded in a horizontal format, as vertical videos often fail to capture complete desktop screens, limiting their utility.
    \item \textbf{Recency.} Videos must be no older than five years to ensure that the user interfaces depicted remain relevant and applicable.
\end{enumerate}

These criteria ensure that selected videos are suitable for detailed analysis and processing in research contexts.

\section{\screenfilter{} Details}
\label{app:screenfilterdetails}

\screenfilter{} is trained on 15,000 synthetic images generated by compositing cursor sprites onto GUI screenshots from the GUIEnv~\citep{chen2025guicoursegeneralvisionlanguage} dataset. To enhance its generalization across different platforms, we incorporate various cursor patterns from both Windows and macOS. On the held-out test set, \screenfilter{} achieves an F1 score of 89.58\%, with 90.64\% precision and 88.54\% recall, demonstrating its effectiveness in accurately separating computer-use content from unrelated material.

For video processing, \screenfilter{} operates at 1-2 frames per second to balance both accuracy and efficiency. The model retains segments where at least 80\% of the frames contain a cursor for a minimum of 6 seconds, with a 2-second temporal smoothing gap to merge frames. This design allows \screenfilter{} to process approximately 840 hours of video per GPU-day, facilitating large-scale filtering.

\section{Dense Event Detection}
\label{app:denseeventdetection}

Our dense detector is trained on \textbf{154 hours} of screen-capture video paired with raw interaction logs from \emph{OpenCUA} \citep{wang2025opencuaopenfoundationscomputeruse}. The logs contain complete demonstrations with precisely timestamped GUI interactions (mouse and keyboard). We convert these logs into prompt-free temporal grounding supervision by mapping low-level events to our action taxonomy, merging short consecutive micro-events into typed spans with start and end timestamps, and discarding segments without actionable GUI operations.

\begin{table}[t]
\centering
\small
\caption{Event distribution in the training data (154 hours).}
\label{tab:dense-stats}
\begin{tabular}{lr}
\toprule
\textbf{Action type} & \textbf{Count} \\
\midrule
click         & 410{,}101 \\
key           & 138{,}660 \\
write         & 80{,}749  \\
scroll        & 46{,}597  \\
moveTo        & 32{,}840  \\
dragTo        & 32{,}840  \\
doubleClick   & 14{,}241  \\
rightClick    & 7{,}451   \\
hscroll       & 3{,}411   \\
hotkey        & 2{,}570   \\
tripleClick   & 2{,}428   \\
middleClick   & 57        \\
\midrule
\textbf{Total} & \textbf{771{,}945} \\
\bottomrule
\end{tabular}
\end{table}

\begin{figure}[t]
  \centering
  \includegraphics[width=0.6\linewidth]{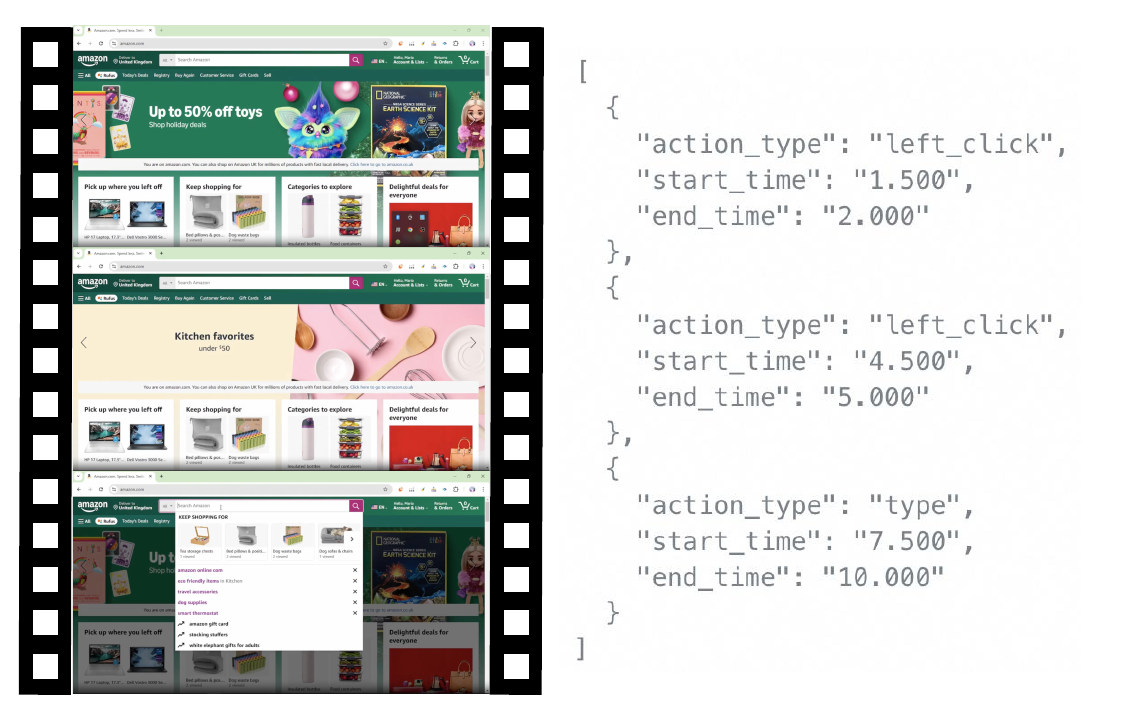}
  \caption{Example training sample for the dense event detector.}
  \label{fig:cg_idm_data}
\end{figure}

For training-set preparation, we downsample videos to 4\,fps, segment them into non-overlapping 10\,s clips, and align the interaction logs within each clip to obtain typed spans with start and end timestamps. We adopt a temporal patch size of 2 frames for modeling efficiency. Label names are normalized to our action taxonomy. We visualize the data sample in Figure \ref{fig:cg_idm_data}

\begin{figure}[t]
  \centering
  \begin{minipage}[t]{0.48\linewidth}
    \centering
    \vspace{0pt}
    \includegraphics[width=\linewidth]{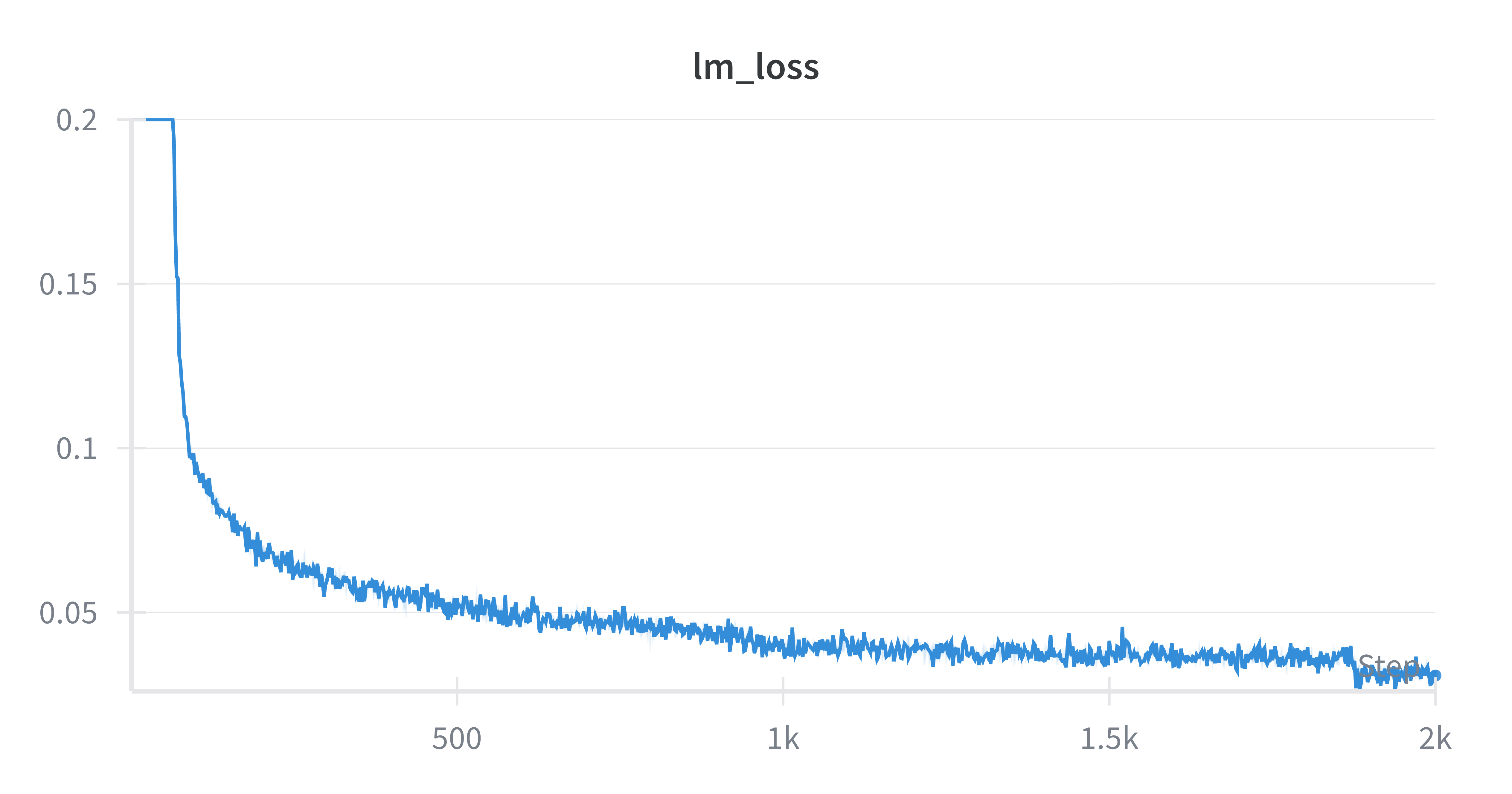}
  \end{minipage}\hfill
  \begin{minipage}[t]{0.48\linewidth}
    \centering
    \vspace{0pt}
    \small
    \begin{tabular}{ll}
    \toprule
    Framework & Megatron-LM \\
    Hardware & 32$\times$ H100 GPUs \\
    Tensor parallelism & TP = 4 \\
    Pipeline parallelism & PP = 1 \\
    Global batch size & 256 \\
    Training iterations & 2000 \\
    LR decay iterations & 2000 \\
    Wall-clock time & $\sim$15 h \\
    \bottomrule
    \end{tabular}
  \end{minipage}
  \caption{Dense event detector: training loss (left) and training configuration (right).}
  \label{fig:cgidm_overview}
\end{figure}

\medskip
\noindent\textbf{Model training.}
We perform full-parameter supervised fine-tuning of Qwen2.5-VL-7B-Instruct. The training configuration and loss curve are shown side-by-side in Figure~\ref{fig:cgidm_overview}.

\section{Action Identification}
\label{app:actionidentification}
\noindent\textbf{Training data.}

\begin{figure}[t]
  \centering
  \includegraphics[width=0.6\linewidth]{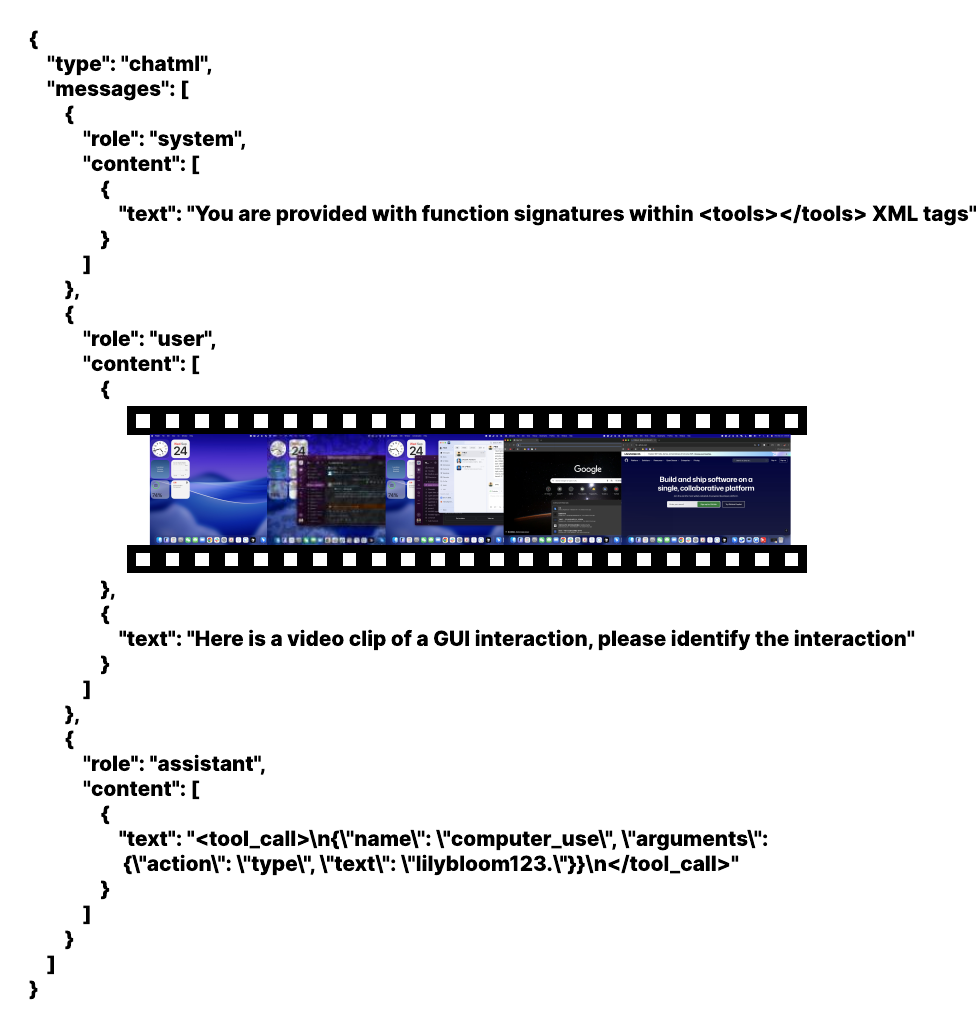}
  \caption{Example training sample for the action parametrization model.}
  \label{fig:fg_idm_data}
\end{figure}

Our dense detector is trained on 512{,}000 screen-capture clips paired with raw interaction logs from \emph{OpenCUA} \citep{wang2025opencuaopenfoundationscomputeruse}. To preprocess action segments, we adopt a \emph{dynamic frame-rate} policy that caps frames per clip at 20 while preserving short, fast interactions. For a segment of duration $\Delta t$ (seconds), we set
\[
f=\min\{30,\ \max\{4,\ \lfloor 20/\Delta t \rfloor\}\},
\]
then sample frames uniformly within $[t_k^{\mathrm{s}},t_k^{\mathrm{e}}]$. This yields, for example, $f{=}30$ for brief clicks/scrolls ($\Delta t\!\approx\!0.5$\,s, $\approx15$ frames), $f{\approx}20$ for $\Delta t\!\approx\!1.0$\,s (20 frames), and $f{=}4$ for extended typing segments ($\Delta t\!\approx\!5$\,s, 20 frames).
We visualize the data sample in Figure

% \begin{minipage}{\textwidth}
\section{Inner Monologue Generation}
\label{app:innermonologuegeneration}

\noindent\textbf{Prompt Content.}
\begin{promptbox}
\textbf{Inputs:}\\
Action type: $a_k$ \quad Parameters: $\pi_k$ \\
Before/after keyframes: $I^{\mathrm{pre}}_k$, $I^{\mathrm{post}}_k$ \\
ASR windows: $[-60\text{s},0]$, $[t_k^{\mathrm{s}},t_k^{\mathrm{e}}]$, $[0,60\text{s}]$

\smallskip
\textbf{Instruction (to model):}\\
You are generating inner-monologue annotations for a dataset of GUI agent trajectories built from in-the-wild screen recordings.

\smallskip
\textbf{End-to-end setting.}
\begin{itemize}[nosep]
  \item Source: real GUI screen recordings from the wild.
  \item Extraction: each GUI interaction (an action) is automatically detected from video/audio.
  \item For every detected action, you receive three kinds of evidence:
  \begin{itemize}
    \item \textit{Action details}: \{action\_type\} and \{action\_content\}.\\
    action\_content may contain: coordinates (absolute or normalized) and/or a bbox; typed text; pressed keys; scroll amount/direction; drag start/end; and similar specifics.
    \item \textit{Keyframes}: a start screenshot and, if available, an end screenshot right after the action executes.
    \item \textit{Surrounding transcripts}: short snippets of narration or speech immediately before, during, and after the action.
    \item \textit{Action validation (optional)}: a brief validator description summarizing what occurred.
  \end{itemize}
\end{itemize}

\smallskip
\textbf{Your task.} For each action, output \textbf{exactly one} JSON object with two fields: \textit{action\_description} and \textit{thought}.

\smallskip
\textbf{Field definitions (strict).}
\begin{itemize}[nosep]
  \item \textbf{action\_description}: a concise natural-language description of \emph{what} I do in the UI at this step. Name the target UI element if inferable (button, menu, tab, field); otherwise describe by role/label/relative position. Mention the immediate visible outcome only if it is clearly observable. \textbf{Forbidden}: raw coordinates, code, function/method names, automation tokens, key–value argument lists.
  \item \textbf{thought}: my first-person inner monologue (4–8 sentences) as the demonstrator (use “I”, “me”, “my”). Provide substantive reasoning. Include: (a) what I aim to accomplish and why now; (b) how the speech context informs my intent (weave naturally); (c) a brief summary of what likely changes from start to end if both frames exist; (d) a short breakdown of the atomic actions in this step (e.g., type + press) and why each is needed; (e) what I expect to verify or do next. Prefer present tense when natural.
\end{itemize}

\smallskip
\textbf{General rules.}
\begin{itemize}[nosep]
  \item The thought must be in first person; never switch to third person.
  \item Evidence priority: prefer visual evidence from start/end keyframes; treat speech as a weak hint for \emph{why}. If they conflict, prefer visuals.
  \item Weave evidence naturally without naming “transcripts” or “frames.”
  \item For coordinate-based actions, a red hollow circle may mark the interaction point; \textbf{do not mention} the marker, describe the target element instead.
  \item If only a start keyframe is available, focus on intent; if an end keyframe exists, you may include the immediate visible result.
  \item When a step bundles multiple atomic actions, reason across them as one coherent operation.
  \item Keep \textit{action\_description} concise; let \textit{thought} carry the details; avoid hedging and boilerplate.
  \item \textbf{Output format}: exactly one valid JSON object with only \textit{action\_description} and \textit{thought}; no extra keys or commentary.
\end{itemize}

\smallskip
\textbf{Output:}\\
$r_k$: inner-monologue JSON with fields \textit{action\_description} and \textit{thought}.
\end{promptbox}
% \end{minipage}

\section{Computer Use Agent Training}
\label{app:trainingstrategy}
\begin{figure}[t]
  \centering
  \begin{minipage}{0.4\linewidth}
    \centering
    \includegraphics[width=\linewidth]{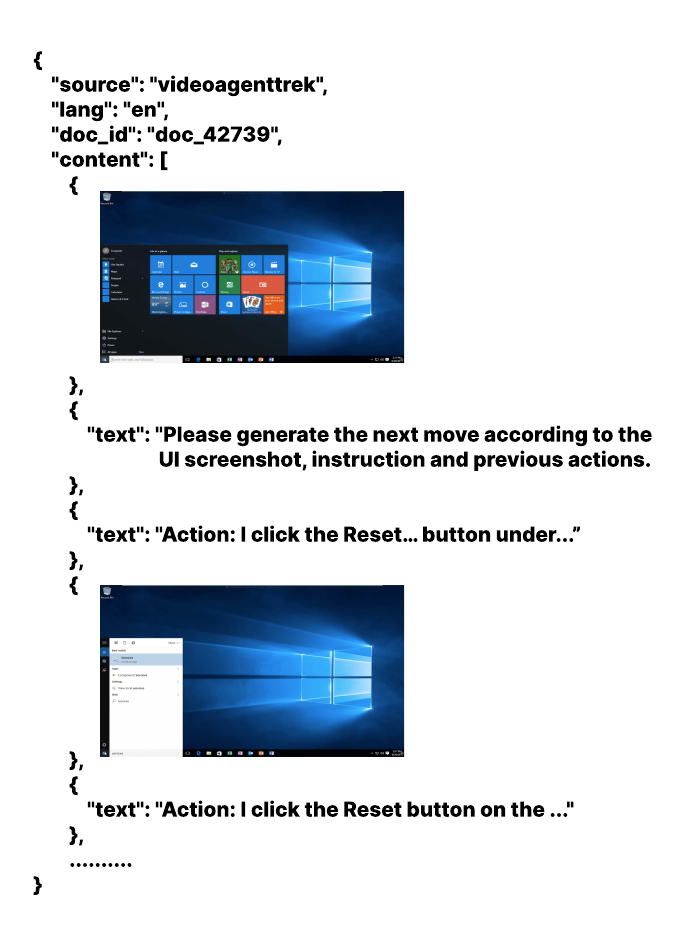}\\[2pt]
    \small (a) Stage-1 training
  \end{minipage}\hfill
  \begin{minipage}{0.4\linewidth}
    \centering
    \includegraphics[width=\linewidth]{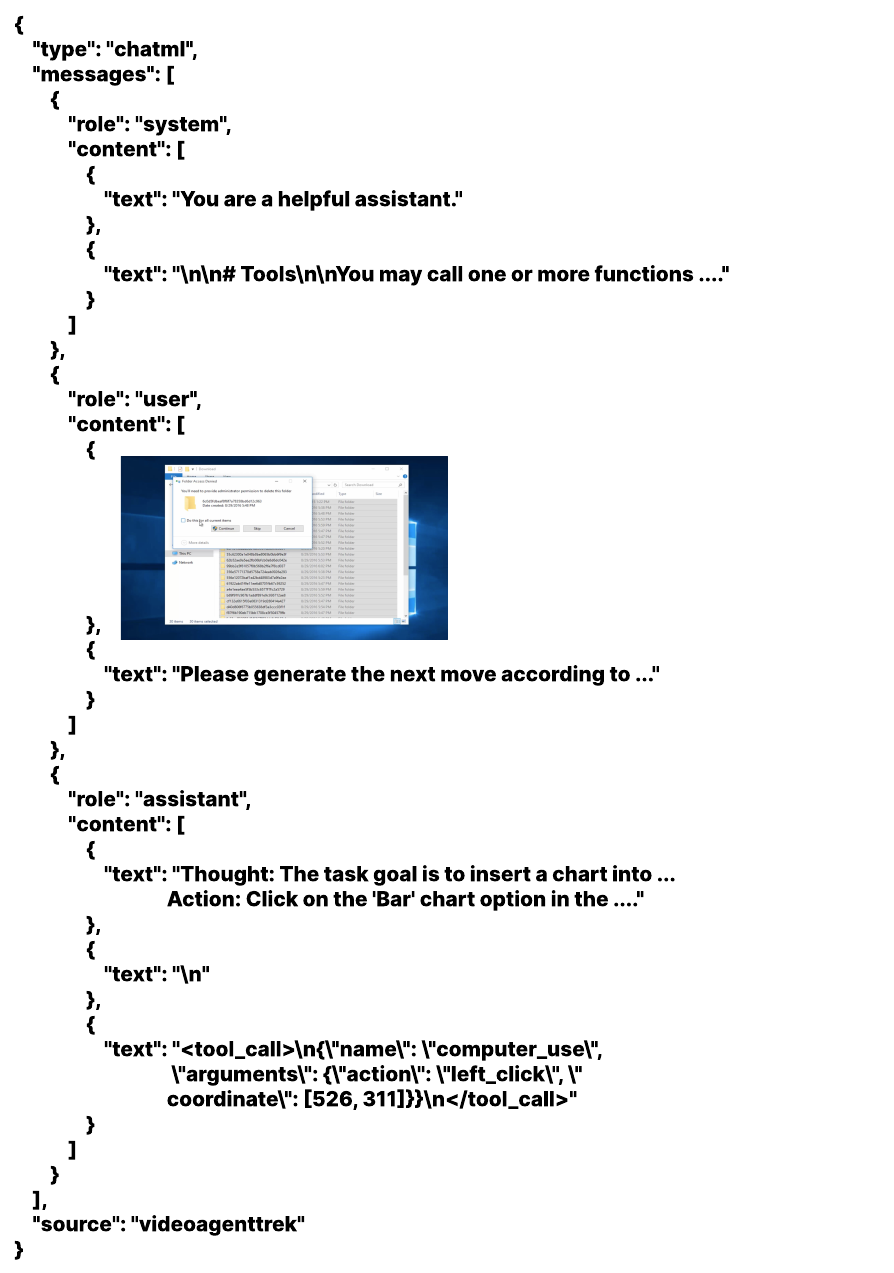}\\[2pt]
    \small (b) Stage-2 training
  \end{minipage}
  \caption{Computer Use Agent Training Data (a) Stage-1 training, (b) Stage-2 training.}
  \label{fig:computerusetrainingdata}
\end{figure}

\noindent\textbf{Training Data.}
we visualize the training data samples in stage-1 and stage-2 training in Figure~\ref{fig:computerusetrainingdata}.

\begin{figure}[t]
  \centering
  \begin{minipage}[t]{0.48\linewidth}
    \centering
    \vspace{0pt}
    \includegraphics[width=\linewidth]{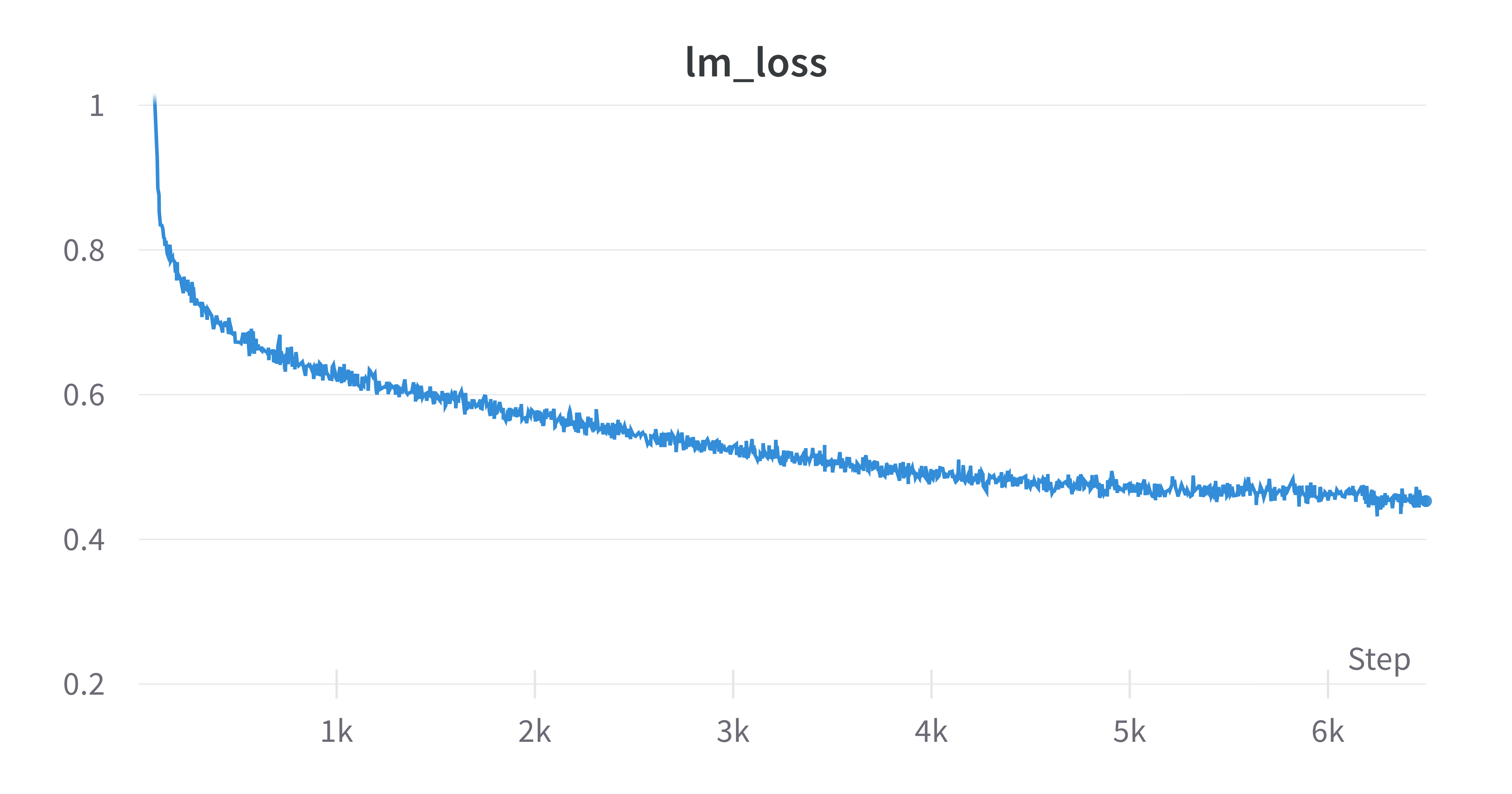}
  \end{minipage}\hfill
  \begin{minipage}[t]{0.48\linewidth}
    \centering
    \vspace{0pt}
    \small
    \begin{tabular}{ll}
    \toprule
    Framework & Megatron-LM \\
    Hardware & 32$\times$ H100 GPUs \\
    Tensor parallelism & TP = 4 \\
    Pipeline parallelism & PP = 1 \\
    Global batch size & 512 \\
    Training iterations & 6500 \\
    LR decay iterations & 6500 \\
    Wall-clock time & $\sim$60 h \\
    \bottomrule
    \end{tabular}
  \end{minipage}
  \caption{CUA Stage-1 Training: training loss (left) and training configuration (right).}
  \label{fig:cuatrainistage1}
\end{figure}

\medskip
\noindent\textbf{Stage-1 training.}
We perform full-parameter continue-pretraining Qwen2.5-VL-7B-Instruct. The training configuration and loss curve are shown side-by-side in Figure~\ref{fig:cuatrainistage1}.

\begin{figure}[t]
  \centering
  \begin{minipage}[t]{0.48\linewidth}
    \centering
    \vspace{0pt}
    \includegraphics[width=\linewidth]{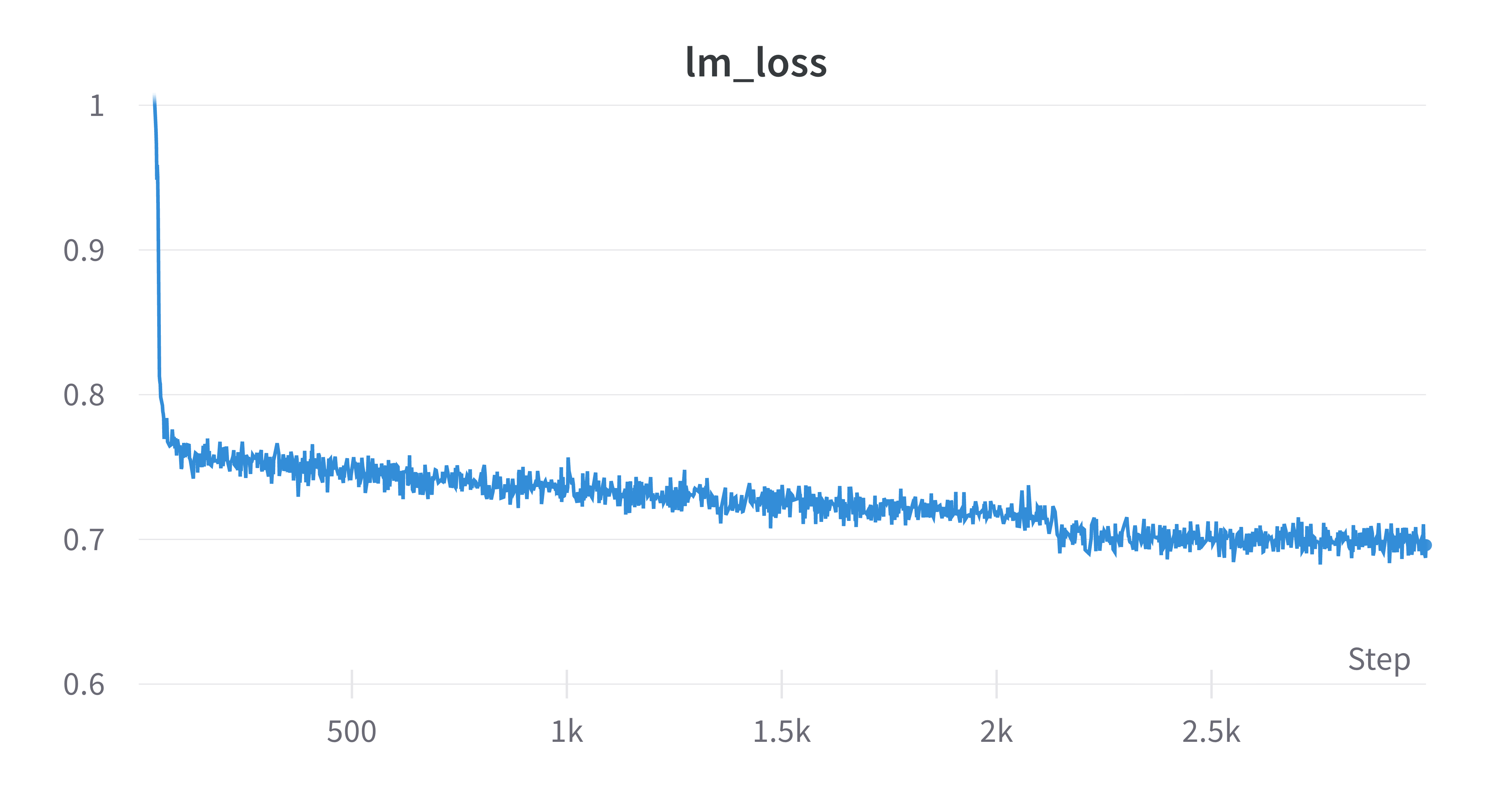}
  \end{minipage}\hfill
  \begin{minipage}[t]{0.48\linewidth}
    \centering
    \vspace{0pt}
    \small
    \begin{tabular}{ll}
    \toprule
    Framework & Megatron-LM \\
    Hardware & 64$\times$ H100 GPUs \\
    Tensor parallelism & TP = 4 \\
    Pipeline parallelism & PP = 1 \\
    Global batch size & 512 \\
    Training iterations & 3000 \\
    LR decay iterations & 3000 \\
    Wall-clock time & $\sim$16 h \\
    \bottomrule
    \end{tabular}
  \end{minipage}
  \caption{CUA Stage-2 Training: training loss (left) and training configuration (right).}
  \label{fig:cuatrainistage2}
\end{figure}

\noindent\textbf{Stage-2 training.}
We continue a full-parameter supervised fine-tuning on the stage-1-trained checkpoint.  The training configuration and loss curve are shown side-by-side in Figure~\ref{fig:cuatrainistage2}.

\section{Computer Use Agent Evaluation}
\label{app:computeruseagentevaluation}
% Preamble: \usepackage{booktabs}\usepackage{multirow}\usepackage{graphicx}

\begin{table}[t]
\centering
\scriptsize
\setlength{\tabcolsep}{4pt}
\resizebox{\linewidth}{!}{%
\begin{tabular}{llcccccccccccc}
\toprule
\textbf{Model} & \textbf{Eval} & \textbf{Task SR (\%)} &
\textbf{calc/46} & \textbf{chrome/46} & \textbf{gimp/26} &
\textbf{vscode/23} & \textbf{writer/23} & \textbf{tbird/15} &
\textbf{os/24} & \textbf{impress/47} & \textbf{workflow/92} & \textbf{vlc/17} \\
\midrule
\multirow{4}{*}{stage2 only}
  & turn1 (20) & 9.42 & 1 & 8 & 3 & 6 & 2 & 3 & 2 & 3 & 5 & 1 \\
  & turn2 (20) & 9.13 & 2 & 8 & 1 & 6 & 2 & 2 & 2 & 3 & 5 & 2 \\
  & turn3 (20) & 9.42 & 1 & 8 & 2 & 6 & 2 & 2 & 2 & 4 & 5 & 2 \\
  & turn4 (50) & 9.27 & 1 & 12 & 2 & 3 & 2 & 2 & 2 & 4 & 4 & 1 \\
\midrule
\multirow{4}{*}{stage1 + stage2}
  & turn1 (20) & 14.68 & 2 & 13 & 2 & 6 & 5 & 6 & 4 & 6 & 7 & 2 \\
  & turn2 (20) & 13.57 & 2 & 13 & 2 & 5 & 5 & 6 & 3 & 6 & 5 & 2 \\
  & turn3 (20) & 14.13 & 2 & 12 & 2 & 7 & 7 & 6 & 3 & 5 & 5 & 2 \\
  & turn4 (50) & \textbf{15.78} & 1 & \textbf{15} & 2 & 6 & 6 & 6 & 4 & 6 & \textbf{8} & 3 \\
\bottomrule
\end{tabular}}
\caption{OSWorld-Verified full results. Counts indicate solved tasks per application bucket (denominators shown in headers).}
\label{tab:osworld_verified_results}
\end{table}
\begin{table}[t]
\centering
\scriptsize
\setlength{\tabcolsep}{5pt}
\resizebox{\linewidth}{!}{%
\begin{tabular}{lccccccccccc}
\toprule
\textbf{Model} & \textbf{Step SR} & \textbf{click} & \textbf{write} & \textbf{press} & \textbf{scroll} & \textbf{moveTo} & \textbf{dragTo} & \textbf{hotkey} & \textbf{dbClick} & \textbf{rClick} & \textbf{terminate} \\
\midrule
base         & 0.385 & 0.402 & 0.605 & 0.286 & 0.615 & 0.189 & 0.000 & 0.250 & 0.000 & 0.000 & 0.188 \\
stage2 only  & 0.641 & 0.671 & 0.719 & ---   & 0.500 & 0.300 & 0.145 & 0.484 & 0.526 & 0.214 & 0.588 \\
stage1 + 2   & \textbf{0.693} & \textbf{0.767} & \textbf{0.733} & \textbf{0.441} & 0.600 & \textbf{0.502} & \textbf{0.264} & \textbf{0.562} & \textbf{0.650} & \textbf{0.417} & 0.237 \\
\bottomrule
\end{tabular}}
\caption{AgentNet Bench: step success rate (overall and per action type). “---” indicates the metric was not applicable/recorded.}
\label{tab:agentnet_bench_results}
\end{table}
\noindent\textbf{Evaluation Setting.}
We follow the OSWorld-Verified protocol: the agent interacts with a live desktop given a natural-language instruction and the full history of prior states and actions. At each step, the policy conditions on the instruction and a bounded visual context of up to five recent screenshots (FIFO window) together with the action/rationale history, then emits the next action. For the 20-step budget, we conduct three independent runs per model and report the average Task SR. For the 50-step budget, we perform a single run. All models use identical inference settings and action executors; no manual interventions are allowed during evaluation.

\noindent\textbf{AgentNet Bench summary.}
Overall step SR rises from 0.385 (base) to 0.641 with SFT-only and to \textbf{0.693} with Stage~1\,+\,Stage~2. These trends suggest that video pretraining notably strengthens grounding and multi-action control, especially for less frequent or harder motor primitives.

\noindent\textbf{OSWorld-Verified summary.}
Table~\ref{tab:osworld_verified_results} reports task success across turns and step budgets. With SFT-only (\emph{stage2 only}), Task SR hovers around 9.1–9.4\% at 20 steps and shows no improvement at 50 steps (9.27\%), indicating limited ability to leverage longer budgets. Adding VideoAgentTrek pretraining (\emph{stage1 + stage2}) raises Task SR to 13.6–14.7\% at 20 steps and further to \textbf{15.78\%} at 50 steps. Per-domain counts improve most for \emph{chrome/46} (up to \textbf{15} solved) and \emph{workflow/92} (up to \textbf{8} solved), with steady gains in \emph{os/24} and authoring apps (writer, impress). Across three 20-step runs, variance is modest, suggesting stable benefits from Stage~1. Overall, the results show that large-scale video pretraining yields higher step quality and makes the agent budget-sensitive—able to convert extra steps into additional task completions.

\newpage

\section{VideoAgentTrek Data Analysis}
\label{app:videoagenttrekdataanalysis}

\noindent\textbf{Resolution and scale.}
We downloaded 55{,}603 screen-capture videos (about 10{,}000 hours) from 50{+} channels. The corpus is predominantly clear: 97\% are 720p or higher (Table~\ref{tab:vatt-res}). Most videos are minutes long, providing sustained interactions suitable for dense detection and action identification.

\begin{table}[h]
\centering
\small
\begin{tabular}{lr}
\toprule
\textbf{Resolution bucket} & \textbf{Count} \\
\midrule
High (1080p+) & 2{,}322 \\
Standard (720p--1080p) & 49{,}589 \\
Low ($<$720p) & 1{,}464 \\
\bottomrule
\end{tabular}
\caption{Resolution distribution of downloaded videos.}
\label{tab:vatt-res}
\end{table}

\noindent\textbf{Title/description-based content classification.}
To quickly audit topical relevance at scale, we apply a lightweight classifier to each video’s title and brief description.
\begin{itemize}[left=0em, labelsep=1em, itemsep=0.5em]
  \item \textbf{Labels.} 
  \begin{itemize}[left=0.52em, labelsep=1em, itemsep=0.5em]
  \item \textbf{A\_tutorial}: hands-on screen tutorials.
  \begin{itemize}
    \item \textit{Include}: step-by-step demonstrations, cursor-driven walkthroughs, “how to …” tasks; frequent UI focus changes; imperative phrasing in titles (“Create…”, “Install…”, “Fix…”).
    \item \textit{Exclude}: talk-style narrations with few concrete on-screen actions; marketing teasers without real steps.
    \item \textit{Signals}: verbs tied to UI operations (open, click, type), timestamps/chapters per step, tool/app names plus action verbs.
  \end{itemize}

  \item \textbf{B\_background}: expository background with incidental screen use.
  \begin{itemize}
    \item \textit{Include}: market share reports, product overviews, concept explainers where the desktop appears only as a backdrop.
    \item \textit{Exclude}: segments that actually show multi-step operations (move to A\_tutorial).
    \item \textit{Signals}: nouns like “overview, history, comparison, review,” charts/stats in title/description, little or no cursor interaction.
  \end{itemize}

  \item \textbf{C\_tech\_talk}: talks or presentations with slides.
  \begin{itemize}
    \item \textit{Include}: conference talks, webinars, lectures; slide navigation with limited live demos.
    \item \textit{Exclude}: talks that transition into substantial live step-by-step demos (then split or relabel A\_tutorial).
    \item \textit{Signals}: “keynote, webinar, seminar, lecture,” speaker names/affiliations, slide thumbnails.
  \end{itemize}

  \item \textbf{D\_unrelated}: off-topic for computer-use learning.
  \begin{itemize}
    \item \textit{Include}: content where a screen appears but no actionable computer-use task is taught (e.g., pure entertainment, face-cam only).
    \item \textit{Exclude}: any clip with consistent stepwise UI operations (move to A\_tutorial).
    \item \textit{Signals}: lifestyle/vlog tags, gameplay without UI instruction, no app/task keywords.
  \end{itemize}
\end{itemize}
  \item \textbf{Procedure.} Single-pass GPT-4.1 classification with a short instruction to choose exactly one of the four labels given the title and short description; no transcript or frames are used. We use the result only for corpus auditing, tag mining, and optional down-weighting in later filters, not as a hard accept/reject gate.
  \item \textbf{Limitations.} Metadata-only classification can mislabel borderline cases (e.g., talks that include substantial demos). Final training sets are still screened by cursor gating, license/PII checks, and downstream detectors.
\end{itemize}

\noindent\textbf{Distribution.}
Class counts and shares are shown in Table~\ref{tab:l1_content_form_table}. The majority are tutorials, indicating strong alignment with our target use case.

\begin{table}[h]
\centering
\begin{minipage}{0.48\textwidth} % Adjust the width to fit both tables
\centering
\small
\begin{tabular}{lrr}
\toprule
\textbf{Label} & \textbf{Count} & \textbf{Share} \\
\midrule
A\_tutorial   & 38{,}700 & 69.6\% \\
B\_background & 12{,}900 & 23.2\% \\
C\_tech\_talk & 2{,}391  & 4.3\% \\
D\_unrelated  & 1{,}612  & 2.9\% \\
\midrule
\textbf{Total} & \textbf{55{,}603} & \textbf{100\%} \\
\bottomrule
\end{tabular}
\caption{distribution from title/description classification.}
\label{tab:l1_content_form_table}
\end{minipage}%
\hfill
\begin{minipage}{0.48\textwidth} % Adjust the width to fit both tables
\centering
\small
\begin{tabular}{lrr}
\toprule
\textbf{Action type} & \textbf{Count} & \textbf{Share (\%)} \\
\midrule
left\_click   & 1,037,617 & 67.1 \\
type          &   214,816 & 13.9 \\
key           &   145,860 &  9.4 \\
scroll        &   111,203 &  7.2 \\
right\_click  &    24,111 &  1.6 \\
double\_click &    11,848 &  0.8 \\
mouse\_move   &     8,441 &  0.1 \\
drag   &     6,372 &  0.1 \\
hscroll       &       196 &  0.0 \\
\midrule
\textbf{Total} & \textbf{1,547,092} & \textbf{100.0} \\
\bottomrule
\end{tabular}
\caption{Action distribution in the VideoAgentTrek agentic dataset.}
\label{tab:vatt-actions}
\end{minipage}
\end{table}

\noindent\textbf{Action distribution.}
Table~\ref{tab:vatt-actions} summarizes action counts in the VideoAgentTrek agentic data.

\noindent\textbf{Cross-dataset comparison.}
We summarize reported average step counts (and task counts when available) for common CUA datasets and include our corpus for context.

\begin{table}[h]
\centering
\begin{minipage}{0.55\textwidth}
\centering
\small
\setlength{\tabcolsep}{8pt}
\begin{tabular}{lrr}
\toprule
\textbf{Dataset} & \textbf{Tasks} & \textbf{Avg.\ Step} \\
\midrule
AndroidControl~\citep{li2024effectsdatascaleui}   & 15{,}283 & 5.5 \\
OS-Genesis~\citep{sun2025osgenesisautomatingguiagent}             & 2{,}451  & 6.4 \\
AgentTrek~\citep{xu2025agenttrekagenttrajectorysynthesis}              & 10{,}398 & 12.1 \\
Mind2Web~\citep{deng2023mind2webgeneralistagentweb}                & 2{,}350  & 7.3 \\
AgentNet~\citep{wang2025opencuaopenfoundationscomputeruse}         & 22{,}625 & 18.6 \\
\midrule
\textbf{VideoAgentTrek}    & \textbf{39{,}000+}      & \textbf{39.25}$^\dagger$ \\
\bottomrule
\end{tabular}
\caption{Average steps across datasets (as reported in their papers). $^\dagger$Estimated from a 5{,}416-trajectory sample in our corpus.}
\label{tab:dataset_step_comparison}
\end{minipage}%
\hfill
\begin{minipage}{0.40\textwidth}
  \centering
  \includegraphics[width=\linewidth]{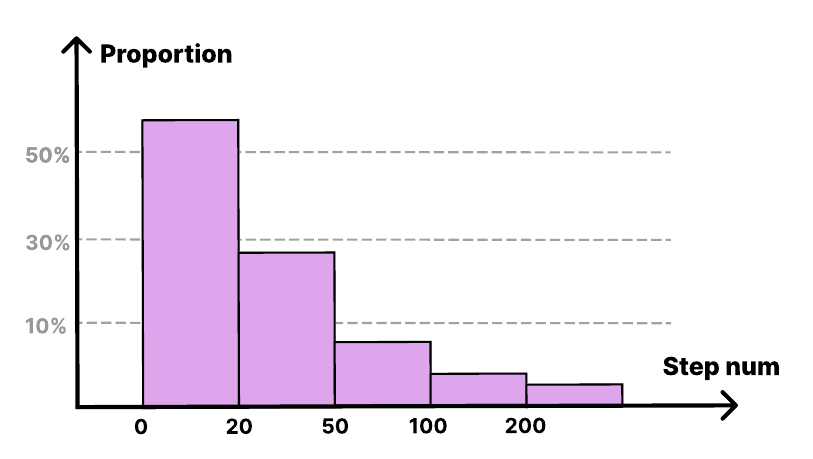}
  \caption{VideoAgentTrek data distribution of step number}
  \label{fig:videoagenttrekdatadistributionofstepnumber}
\end{minipage}
\end{table}

\end{document}